\def\year{2018}\relax
\documentclass[journal,transmag]{IEEEtran}

\usepackage{times}
\usepackage{helvet}
\usepackage{courier}

\usepackage{epsfig}
\usepackage{graphicx}
\usepackage{amsmath}
\usepackage{amssymb}
\usepackage{bm}
\usepackage{mysymbols}
\usepackage{caption}
\usepackage{subcaption}
\usepackage{float}
\usepackage{multirow}
\usepackage{makecell}
\usepackage{wasysym}
\usepackage{epstopdf}
\usepackage{float}

\usepackage[hidelinks,bookmarks=false]{hyperref}

\begin{document}

\title{Thinking Outside the Pool:\\Active Training Image Creation for Relative Attributes}

\author{Aron Yu\\
University of Texas at Austin\\
{\tt\small aron.yu@utexas.edu}
\and
Kristen Grauman\\
Facebook AI Research\\
{\tt \small grauman@fb.com}{$^\ast$}
\thanks{\emph{$^\ast$On leave from UT Austin, grauman@cs.utexas.edu}}
}

\maketitle

\begin{abstract}
  Current wisdom suggests more labeled image data is always better, and obtaining labels is the bottleneck.  Yet curating a pool of sufficiently diverse and informative images is itself a challenge.  In particular, training image curation is problematic for fine-grained attributes, where the subtle visual differences of interest may be rare within traditional image sources.  We propose an active image generation approach to address this issue.  The main idea is to jointly learn the attribute ranking task while also learning to generate novel realistic image samples that will benefit that task.  We introduce an end-to-end framework that dynamically ``imagines'' image pairs that would confuse the current model, presents them to human annotators for labeling, then improves the predictive model with the new examples.  With results on two datasets, we show that by thinking outside the pool of real images, our approach gains generalization accuracy for challenging fine-grained attribute comparisons.
\end{abstract}

\vspace*{0.12in}
\section{Introduction}
\label{sec:intro}
\vspace*{0.05in}

Visual recognition methods are famously data hungry.  Today's deep convolutional neural networks (CNNs) achieve excellent results on various challenging tasks, and one critical ingredient is access to large manually annotated datasets~\cite{kinetics,imagenet,lfw,aron-cvpr14}.  A common paradigm has emerged where a supervised learning task is defined, the relevant images/videos are scraped from the Web, and crowdworkers are enlisted to annotate them appropriately.  A practical downside to this paradigm, of course, is the expense of getting all those annotations, which can be significant.

However, setting cost concerns aside, we contend that the standard paradigm also quietly suffers a \emph{curation} concern.  The assumption is that the more data we gather for labeling, the better off we will be.  And indeed unlabeled photos are virtually unlimited.
Yet---particularly when relying on Web photos---the visual variety and information content of the images amassed for labeling eventually reach a limit.  The resulting deep network inherits this limit in terms of how well it can generalize.  Importantly, while \emph{active learning} methods can try to prioritize informative instances for labeling, the curation problem remains: existing active selection methods scan the pool of manually curated unlabeled images when choosing which ones to label~\cite{cost-cvpr09,freytag14,sudheendra-ijcv2014,zhao-hcomp2011}.

\begin{figure}[t]
  \centering
  \begin{subfigure}[t]{.47\textwidth}
    \includegraphics[width=\textwidth]{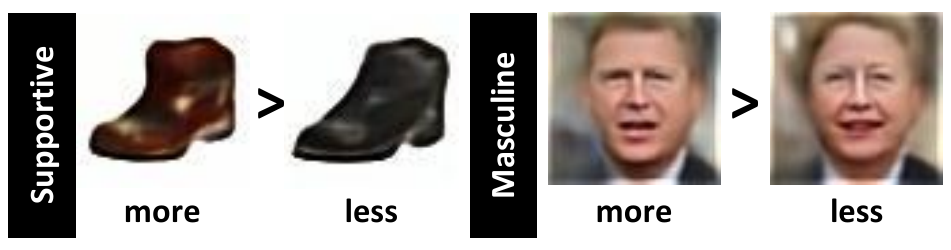}
  \end{subfigure}
  \vspace*{-0.05in}
  \caption{Our approach learns to \emph{actively create} training images for fine-grained attribute comparisons, focusing attention on informative pairs of images unavailable in the pool of real training images.  Here we show two example pairs it generated to improve learning of \emph{supportive} (left) and \emph{masculine} (right).}
  \label{fig:concept}
  \vspace*{0.08in}
\end{figure}

We propose to address the curation challenge with \emph{active image generation}.  The main idea is to jointly learn the target visual task while also learning to generate novel realistic image samples that, once manually labeled, will benefit that task.
We focus specifically on the fine-grained relative attribute task~\cite{relattr11,Sandeep14,Singh16,Souri16,Yang16}: given two images, the system must infer which one more evidently exhibits an attribute (e.g., which is more \emph{furry}, more \emph{red}, more \emph{smiling}). See Figure~\ref{fig:concept}.  The curation problem is particularly pronounced for fine-grained relative attributes.  This is because while at test time an image pair may be arbitrarily close in an attribute, it is difficult to locate extensive training samples showing such subtle differences for all attributes~\cite{aron-iccv17}.  Our idea is for the system to dynamically ``imagine'' image pairs that would confuse the current learning-to-rank model, then present them to human annotators for labeling.

To this end, we introduce an end-to-end deep network consisting of an attribute-conditioned image generator, a ranking network, and a \emph{control} network.  The network is seeded with real labeled image pairs of the form ``image $X$ exhibits attribute $a$ more than image $Y$''.   Then the control network learns to generate latent visual parameters to present to the image generator so as to form image pairs that would be difficult for the ranker.  Thus, rather than simply perturb existing training images (\`a la ``fooling images''~\cite{Nguyen15}), the control module learns a \emph{function} to create novel instances, potentially improving the exploration of the relevant image space.  We train the ranker and controller in an adversarial manner, and we solicit manual labels for the resulting images in batches.\footnote{We use ``batch" here in the active learning sense: a \emph{batch} of additional examples are manually labeled then used to update the predictive model.  This is not to be confused with \emph{(mini)-batches} for training the neural networks.}  As a result, the set of imagined training instances continues to evolve, as does the ranker.

Our main contribution is a new approach to active learning in which pairs of samples are \emph{newly created} so as to best augment a ranker's training data.  Through experiments on two challenging fine-grained attribute datasets of Shoes~\cite{aron-cvpr14,aron-iccv17} and Faces~\cite{lfw,Sandeep14}, we demonstrate the approach's effectiveness.   Active generation focuses attention on novel training images that rapidly improve generalization---even after all available real images and their labels are exhausted.  It outperforms alternative data augmentation strategies, including prior work that generates synthetic training images in a passive manner.

\vspace*{0.1in}
\section{Related Work}
\label{sec:related}
\vspace*{0.1in}

\noindent
\textbf{Attribute Comparisons:}  Relative attributes~\cite{relattr11} model visual properties in a comparative manner, and a variety of applications have been explored in online shopping~\cite{whittle-ijcv}, fashion~\cite{hipster}, biometrics~\cite{nixon-attributes}, and graphical design~\cite{relfonts14}.  Recent work explores novel learning schemes to train relative attributes accurately, including new deep network architectures~\cite{Singh16,Souri16,Yang16}, part discovery~\cite{Sandeep14}, local learning~\cite{aron-cvpr14,aron-iccv15}, and multi-task approaches~\cite{Meng18}.  Our contribution is an approach to actively elicit training examples for ranking, which could facilitate training for many of the above formulations.

\vspace*{0.1in}
\noindent
\textbf{Image Generation:}  Photo-realistic image generation has made steady and exciting process in the recent years, often using generative adversarial networks (GANs)~\cite{Goodfellow14,dcgan,pix2pix,cycgan,stackgan,sagan} or variational auto-encoders (VAEs)~\cite{Gregor15,Kulkarni15,Yan16-eccv}.  Conditional models can synthesize an image based on an input, either a label map~\cite{pix2pix,cycgan} or a latent attribute label~\cite{fader,upchurch,Yan16-eccv}.  We integrate the Attribute2Image~\cite{Yan16-eccv} network as the image generation module in our full pipeline, though similarly equipped conditional models could also be used.  Methods to generate adversarial ``fooling images'' also use a network to automatically perturb an image in a targeted manner~\cite{Baluja17,DeepFool16,Nguyen15}. However, rather than using adversarial generation to understand how features influence a classifier, our goal is to synthesize the very training samples that will strengthen a learned ranker.  Unlike any of the above, we create images for active query synthesis.

\vspace*{0.1in}
\noindent
\textbf{Learning with Synthetic Images:}  The idea of training on synthetic images but testing on real ones has gained traction in recent years.  Several approaches leverage realistic human body rendering engines to learn 3D body pose~\cite{psh,shotton-cvpr2011,surreal}, while others extrapolate from video~\cite{chaoyeh-cvpr2013,ParkRamanan,Khoreva17} or leverage 3D scene~\cite{Zhang17} or object~\cite{saenko-iccv2015,Yang18} models.  Refiner networks that translate between the simulated and real data can help ease the inevitable domain gap~\cite{Shrivastava17}.

Data augmentation (used widely in practice) can be seen as a hybrid approach, where each real training sample is expanded into many samples by applying low-level label-preserving transformations, or ``jitter", like scaling, rotating, mirroring, etc.~\cite{augment}.  Usually the scope of jitter is manually defined, but some recent work explores how to adversarially generate ``hard'' jitter for body pose estimation~\cite{Peng18}, greedily select useful transformations for image classification~\cite{Paulin14}, or actively evolve part-based 3D shapes to learn shape from shading~\cite{Yang18}.  Such methods share our motivation of generating samples where they are most needed, but for very different tasks.  Furthermore, our approach aligns more with active learning than data augmentation: the new synthetic samples can be distant from available real samples, and so they are manually annotated before being used for training.

In the low-shot recognition regime, several recent methods explore creative ways to hallucinate the variability around sparse real examples~\cite{Kwitt16,Dixit17,Hariharan17,aron-iccv17,dreaming,Schwartz18}, typically leveraging the observed inter-sample transformations to guide synthesis in feature-space.  Among them, most relevant to our work is the ``semantic jitter'' approach~\cite{aron-iccv17}, which generates relative attribute training images by altering one latent attribute variable at a time.  However, whereas semantic jitter~\cite{aron-iccv17} uses manually defined heuristics to sample synthetic training images, we show how to dynamically derive the images most valuable to training, via an adversarial control module learned jointly with the attribute ranker.

\vspace*{0.1in}
\noindent
\textbf{Active Learning:}  Active learning has been studied for decades~\cite{Settles10}.  For visual recognition problems, \emph{pool-based} methods are the norm: the learner scans a pool of unlabeled samples and iteratively queries for the label on one or more of them based on uncertainty or expected model influence (e.g.,~\cite{freytag14,sudheendra-ijcv2014,zhao-hcomp2011,cost-cvpr09}).  Active ranking models adapt the concepts of pool-based learning to select pairs for comparison~\cite{Qian14,lucy-cvpr2014}.  Hard negative mining---often used for training object detectors~\cite{pedro-2008,hard-negs}---also focuses the learner's attention on useful samples, though in this case from a pool of already-labeled data.  Rather than display one query image to an annotator, the approach in~\cite{Huijser17} selects a sample from the pool then displays a synthesized image \emph{spectrum} around it in order to elicit feature points likely to be near the true linear decision boundary for image classification.  We do not perform pool-based active selection.  Unlike any of the above, our approach \emph{creates} image samples that (once labeled) should best help the learner, and it does so in tight coordination with the ranking algorithm.

In contrast to pool-based active learning, active \emph{query synthesis} methods request labels on novel samples from a given distribution~\cite{query-halfspace,query-ml,Settles10,Zhu17}.  When the labeler is a person (as opposed to an oracle or experimental outcome), a well known challenge with query synthesis is that the query may turn out to be unanswerable~\cite{baum}.  Perhaps accordingly, there is very limited prior work attempting active query synthesis for image problems, and to our knowledge they are limited to toy cases like MNIST digits~\cite{Zhu17}. Our work capitalizes on the recent advances in image generation discussed above to create photorealistic queries that are most often answerable. Furthermore, rather than sample from an input distribution, our selection approach is discriminative: it optimizes the latent parameters of an image pair that directly affect the current deep ranking model.

\vspace*{0.1in}
\section{Approach}
\label{sec:approach}
\vspace*{0.05in}

We propose an end-to-end framework for attribute-based image comparison through active adversarial image generation.
We refer to our approach as \emph{ATTIC}, for \textbf{A}c\textbf{T}ive \textbf{T}raining \textbf{I}mage \textbf{C}reation.

\begin{figure}[t]
  \centering
  \begin{subfigure}[t]{.48\textwidth}
    \includegraphics[width=\textwidth]{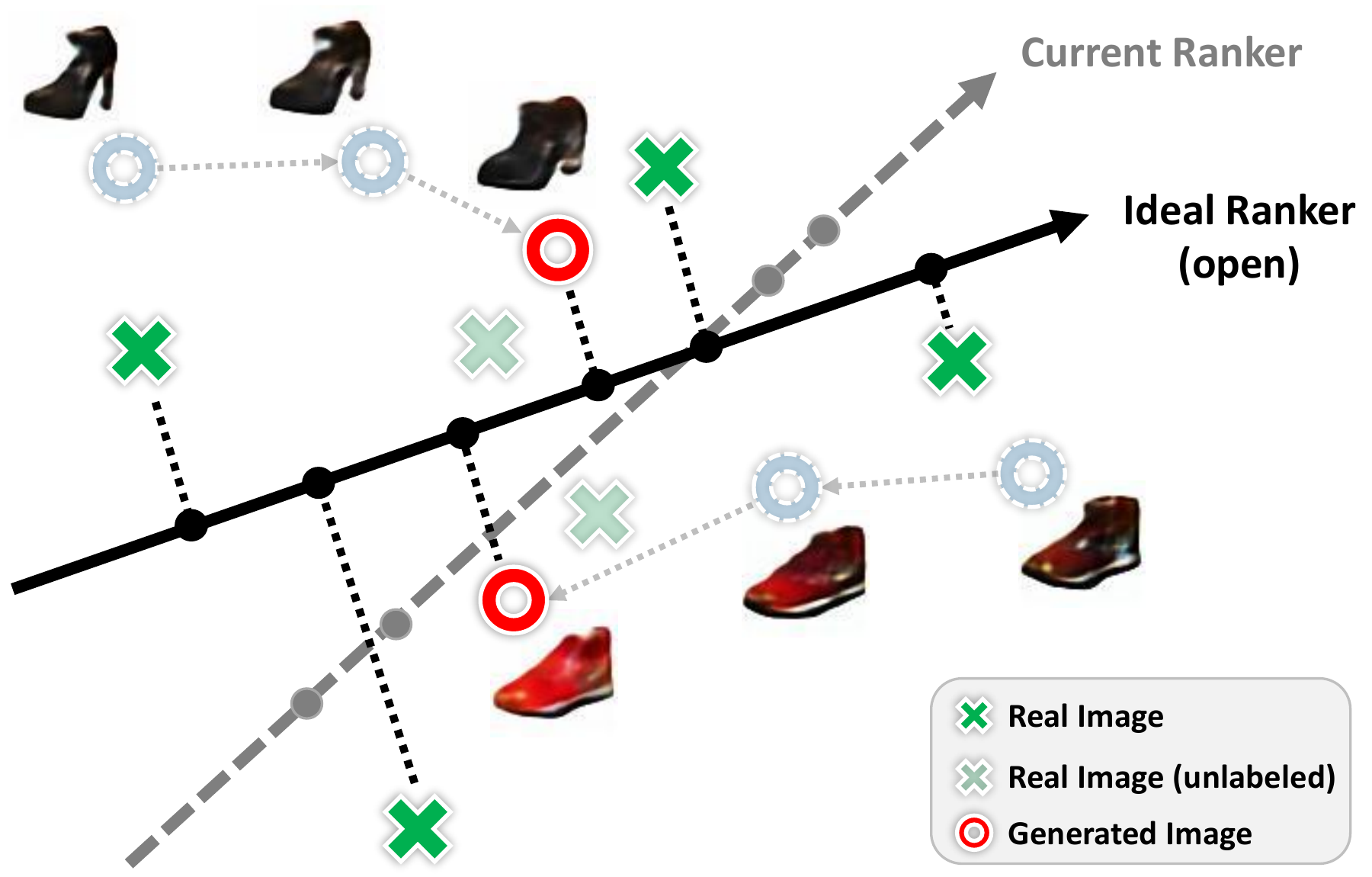}
  \end{subfigure}
  \vspace*{0.05in}
  \caption{Schematic overview of main idea.  Real images (green $\times$'s) are used to train a deep ranking function for the attribute (e.g., the \emph{openness} attribute for shoes). The pool of real images consists of those that are labeled (dark $\times$'s) and those that are unlabeled (faded $\times$'s).  Even with all the real images labeled, the ideal ranking function may be inadequately learned.  Rather than select  other manually curated images for labeling (faded green $\times$'s), ATTIC directly generates useful synthetic training images (red $\ocircle$'s) through an adversarial learning process.  The three shoes along each path of circles represent how ATTIC iteratively evolves the control parameters to obtain the final synthetic image pairs.}
  \label{fig:approach}
  \vspace*{0.15in}
\end{figure}

The goal is to avoid the ``streetlight effect'' of traditional pool-based active learning, where one looks for more training data where one already has it.\footnote{\scriptsize{\texttt{\url{https://en.wikipedia.org/wiki/Streetlight_effect}}}}  Rather than limit training to manually curated real images, as would standard pool-based active learning, ATTIC synthesizes image pairs that will be difficult for the ranker as it has been trained thus far.  See Figure~\ref{fig:approach}.

We first define the relative attributes problem (Sec.~\ref{sec:ranking}).  Then we describe the key modules in our ATTIC network (Sec.~\ref{sec:attic}).  Finally we define our training procedure and the active image creation active loop (Sec.~\ref{sec:training}).


\vspace*{0.05in}
\subsection{Relative Attributes via Ranking}
\label{sec:ranking}
\vspace*{0.05in}

Relative attributes are valuable for inferring fine-grained differences between images in terms of describable natural language attributes.  They have wide application in image search, zero-shot learning, and biometrics, as discussed above.
At the core of the relative attribute comparison task is a learning-to-rank problem~\cite{relattr11,Singh16}.  A ranking function $R_\mathcal{A} : X \rightarrow \mathbb{R}$ for attribute $\mathcal{A}$ (e.g., \emph{comfortable}, \emph{smiling})  maps an image to a real-valued score indicating the strength of the attribute in the image.

Relative attribute rankers are trained using \emph{pairs} of training images.  Let $\mathcal{P}_\mathcal{A} = \{(\bm{x}_i,\bm{x}_j)\}$ be a set of ordered pairs, where humans perceive image $\bm{x}_i$ to have ``more'' of attribute $\mathcal{A}$ than image $\bm{x}_j$.  The goal is to learn a ranking function $R_\mathcal{A}$ that satisfies the relative orderings as much as possible: $\forall (i,j) \in \mathcal{P}_\mathcal{A}: R_\mathcal{A}(\bm{x}_i) > R_\mathcal{A}(\bm{x}_j)$.  Having trained such a function $R_\mathcal{A}$ we can use it to score images for their attribute strength individually, or, similarly, to judge which among a novel pair of images $(\bm{x}_m,\bm{x}_n)$ exhibits the attribute more.  ATTIC trains a network for each attribute $\mathcal{A}$; we drop the subscript below when not needed for clarity.

\vspace*{0.05in}
\subsection{End-to-End Active Training Image Creation}
\label{sec:attic}
\vspace*{0.05in}

Let $\mathcal{P}^R$ be an initial set of real training image pairs used to initialize the ranker $R$ defined above.  Our goal is to improve that ranker by creating synthetic training images $\mathcal{P}^S$, to form a hybrid training set $\mathcal{P} = \mathcal{P}^R \bigcup \mathcal{P}^S$.

\begin{figure*}[t]
  \centering
  \includegraphics[width=0.98\textwidth]{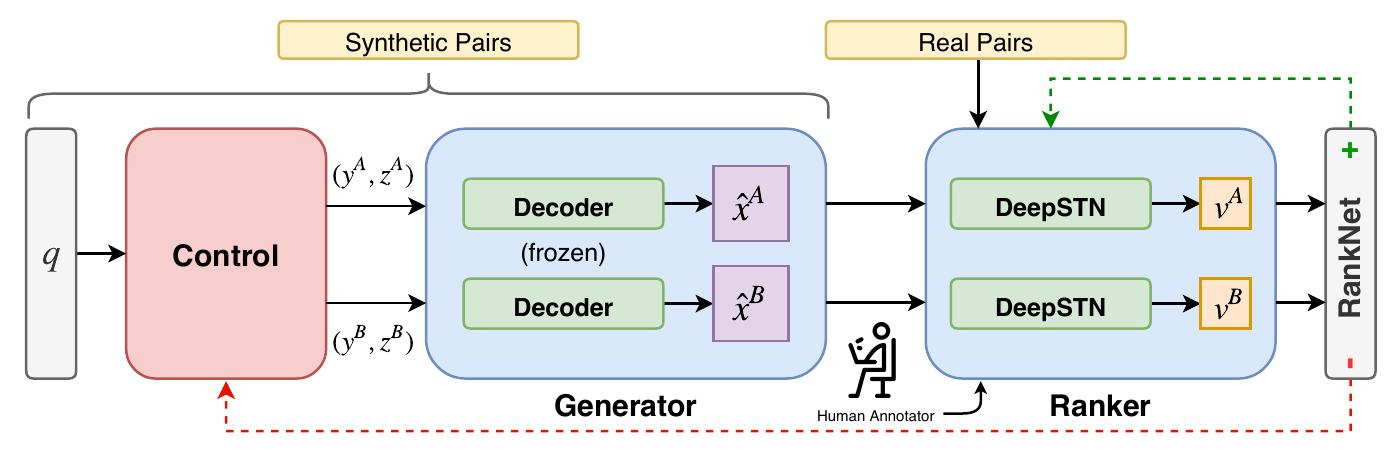}
  \vspace*{-0.05in}
  \caption{Architecture of our proposed end-to-end approach consisting of three primary modules.  The control module first converts the random input $q$ into control parameters $\{(\bm{y}^A,\bm{z}^A),(\bm{y}^B,\bm{z}^B)\}$. Its architecture is detailed further in Figure~\ref{fig:control}.  The generator module then generates a pair of synthetic images $(\hat{\bm{x}}^A,\hat{\bm{x}}^B)$ using these control parameters.  The ranker module finally uses the generated synthetic images (once manually labeled) and the real training images to train the ranking model, outputting their corresponding attribute strength $(v^A,v^B)$.  During training, the ranking loss using the RankNet objective is fed back into the ranker (green dotted line), while the negative ranking loss from the same objective is fed back into the control module (red dotted line).  Note that the decoders within the generator are pre-trained and their parameters are kept frozen throughout training.}
  \label{fig:overview}
  \vspace*{-0.05in}
\end{figure*}

To this end, the proposed end-to-end ATTIC framework consists of three distinct components (Fig.~\ref{fig:overview}): the ranker module, the generator module, and the control module.  Our model performs end-to-end adversarial learning between the ranker and the control modules.  The ranker tries to produce accurate attribute comparisons, while the control module tries to produce \emph{control parameters}---latent image parameters---that will generate difficult image pairs to confuse the current ranker.  By asking human annotators to label those confusing pairs, the ranker is actively improved.   We next discuss the individual modules.


\vspace*{0.15in}
\noindent
\textbf{Ranking Module:}  For the ranking module in ATTIC, we employ the state-of-the-art deep DeepSTN approach~\cite{Singh16}, which integrates a Siamese embedding and a spatial transformer network (STN)~\cite{STN} together with a RankNet ranking loss~\cite{ranknet}.  RankNet handles pairwise outputs in a single differentiable layer using cross-entropy loss.  The rank estimates $(v_i,v_j)$ for images $(\bm{x}_i,\bm{x}_j)$ are mapped to a pairwise posterior probability using a logistic function
\begin{equation}
  p_{ij} = \frac{1}{1 + e^{-(v_i-v_j)}},
  \vspace*{0.05in}
\end{equation}
and the ranking loss is:
\begin{equation}
  \mathcal{L}_{rank} = - \log(p_{ij}).
  \label{eq:ranknet}
  \vspace*{0.05in}
\end{equation}
The ranker jointly learns a CNN image encoder $\phi(\bm{x})$ with a spatial transformer that localizes the informative image regions for judging a particular attribute.  For example, the STN may learn to clue into the mouth region for the \emph{smiling} attribute.   The ranker combines the features of the full image and this region to compute the attribute strength:
\begin{equation}
  v_i = R_\mathcal{A}(\bm{x}_i) = \textrm{RankNet}_\mathcal{A}(\textrm{STN}(\phi(\bm{x}_i))),
  \vspace*{0.05in}
\end{equation}
where $\phi(\bm{x})$ denotes application of one stack in the Siamese embedding.  See~\cite{Singh16} for details.

\vspace*{0.15in}
\noindent
\textbf{Generator Module:}  For the generator module, we use an existing attribute-conditioned image generator, Attribute2Image~\cite{Yan16-eccv}.  Let $\bm{y} \in \mathbb{R}^{N}$ denote an $N$-dimensional attribute vector containing the real-valued strength of each of the $N = |\mathcal{A}|$ attributes for an image $\bm{x}$, and let $\bm{z} \in \mathbb{R}^{M}$ denote an $M$-dimensional latent variable accounting for all other factors affecting the image, like lighting, pose, and background.  The Attribute2Image network uses a Conditional Variational Auto-Encoder (CVAE) to learn a decoder $p_\theta(\hat{\bm{x}} | \bm{y},\bm{z})$ that can generate realistic  synthetic images $\hat{\bm{x}}$ conditioned on $(\bm{y},\bm{z})$.  The parameters $\theta$ are optimized by maximizing the variational lower bound of the log-likelihood $\log p_\theta(\hat{\bm{x}} | \bm{y})$.  The network uses a multilayer perceptron to map the latent factors into their entangled hidden representation, then decodes into image pixels with a coarse-to-fine convolutional decoder.  See~\cite{Yan16-eccv} for more details.

The \emph{attribute-conditioned} aspect of this generator allows us to iteratively refine the generated images in a semantically smooth manner, as we adversarially update its inputs $(\bm{y}, \bm{z})$ with the control module defined next.
We pre-train the generator using $\{(\bm{x}_i,\bm{y}_i)\}$, a disjoint set of training images labeled by their $N$ attribute strength labels.  Subsequently, we take only the decoder part of the model and use it as our generator (see Fig.~\ref{fig:overview}).  We freeze all parameters in the generator during active image creation, since the mapping from latent parameters to pixels is independent of the rank and control learning.


\begin{figure}[t]
  \centering
  \begin{subfigure}[t]{.48\textwidth}
    \includegraphics[width=\textwidth]{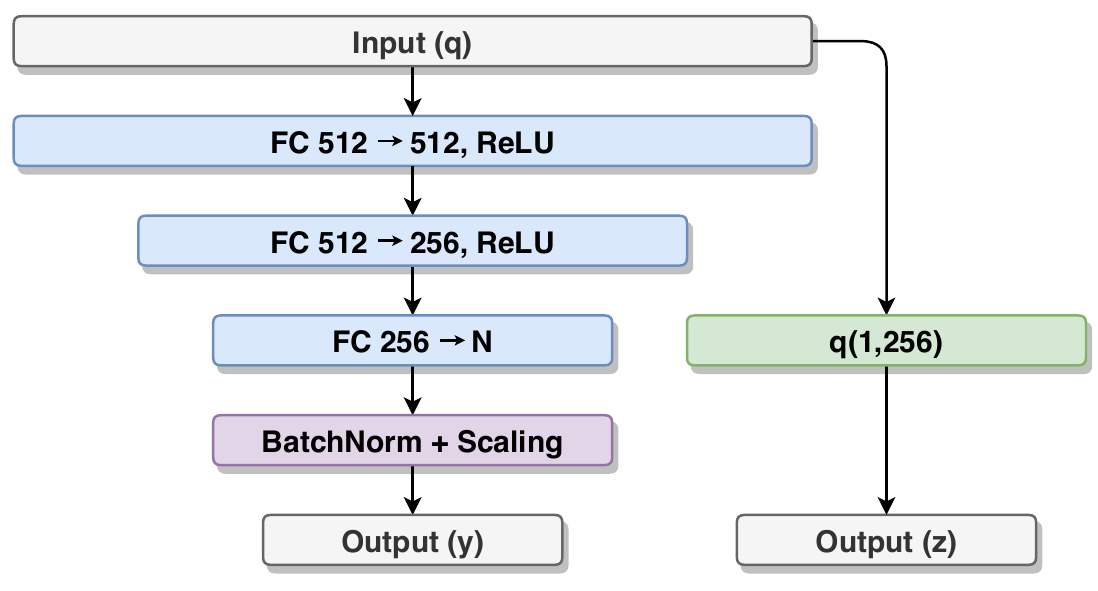}
  \end{subfigure}
  \vspace*{-0.05in}
  \caption{Architecture of the control module.  The model above outputs a single set of control parameters $(\bm{y},\bm{z})$.  Since we generate the synthetic images in pairs, we duplicate the architecture.}
  \label{fig:control}
  \vspace*{0.12in}
\end{figure}

\noindent
\textbf{Control Module:}  As defined thus far, linking together the ranker and generator would aimlessly feed new image sample pairs to the ranker.  Next we define our control module and explain how it learns to feed pairs of intelligently chosen latent parameters to the generator for improving the ranker.

The control module is a neural network that precedes the generator (see Figure~\ref{fig:overview}, left).  Its input is a random seed $\bm{q} \in \mathbb{R}^Q$, sampled from a multivariate Gaussian.  Its output is a pair of \emph{control parameters} $\{(\bm{y}^A,\bm{z}^A), (\bm{y}^B, \bm{z}^B)\}$  for synthetic image generation.  Figure~\ref{fig:control} shows the control architecture. It is duplicated to create two branches feeding to the generator and then the Siamese network in the DeepSTN ranker.

The attribute control variable $\bm{y}$ is formed by passing $\bm{q}$ through a few fully-connected layers, followed by a BatchNorm layer with scaling.  In particular, for the scaling we obtain the scaling parameters from the mean and the standard deviation of the attribute strengths observed from the real training images, then applying them to the normalized $\mathcal{N}(0,1)$ outputs from the BatchNorm layer.  The scaling ensures that the attribute strengths are bounded within a range appropriate for the pre-trained generator.

The latent feature control variable $\bm{z}$, which captures all the non-attribute properties (e.g., pose, illumination), is sampled from a Gaussian.  We simply use half of the entries from $\bm{q}$ for $\bm{z}^A$ and $\bm{z}^B$, respectively.  This Gaussian sample agrees with the original image generator's prior $p(\bm{z})$.

\subsection{Training and Active Image Creation}
\label{sec:training}

Given the three modules, we connect them in sequence to form our active learning network.  The generator and the ranker modules are duplicated for both branches to account for two images in each training pair.  The decoders in the generator module are pre-trained and their parameters are kept frozen.  During training, we optimize the RankNet loss for the ranker module, while at the same time optimizing the negative RankNet loss for the control module:
\begin{equation}
  \mathcal{L}_{control} = -\mathcal{L}_{rank},
  \label{eq:control_loss}
\end{equation}
creating the adversarial effect.  The control module thus learns to produce parameters that generate image pairs that are difficult for the ranker to predict.  This instills an adversarial effect where the control module and the ranker module are competing against each other iteratively during training.  The learning terminates when the ranker converges.

Figure~\ref{fig:spectrum} in the next section shows examples of the \emph{progression} of some synthetic images during the training iterations. ATTIC captures the joint interaction between the attributes and traverses the multi-attribute space to places not occupied by real training images.  

To generate a batch\footnote{Again, here we mean ``batch'' in the active learning sense: a batch of examples to be labeled and then augment the training set.} of synthetic image pairs $\mathcal{P}^S = \{\hat{\bm{x}}^A,\hat{\bm{x}}^B)\}_{i=1}^T$, we sample $T$ vectors $\bm{q}$ and push them through the control and generator.  Then the batch is labeled by annotators, who judge which image shows the attribute more, and the resulting pairs accepted by annotators as valid are added to the hybrid training set $\mathcal{P}$.

The primary novelty of our approach comes from the generation of synthetic image pairs through active query synthesis.  From an active learning perspective, instead of selecting more real image pairs to be labeled based on existing pool-based strategies, our approach aims to directly generate the most beneficial synthetic image pairs (Fig.~\ref{fig:approach}).  Furthermore, instead of choosing $(\bm{y}_i,\bm{z}_i)$ manually when generating the synthetic image pairs, as proposed in Semantic Jitter~\cite{aron-iccv17}, ATTIC automates this selection in a data-driven manner.


\begin{table*}[t]
  \begin{center}\footnotesize
  \renewcommand{\arraystretch}{1.2}
  \begin{tabular}{|c||l||*{10}{c}|}
  \cline{2-12}
  \multicolumn{1}{c|}{} & & \Gape[4pt]{\textbf{Comfort}} & \textbf{Casual} & \textbf{Simple} & \textbf{Sporty} & \textbf{Colorful} & \textbf{Durable} & \textbf{Supportive} & \textbf{Bold} & \textbf{Sleek} & \textbf{Open} \\
  \Xhline{1pt}
  \multirow{5}{*}[0ex]{\rotatebox{90}{\textbf{Normal}}} & Real~\cite{Singh16} & \textbf{84.26} & 88.58 & 88.99 & 88.18 & \textbf{94.10} & 82.83 & 83.96 & 88.25 & 84.35 & \textbf{83.87} \\
  & Real+ & 81.71 & 87.96 & 87.12 & 87.58 & 91.05 & 82.60 & 84.41 & 87.63 & 85.82 & \textbf{83.87} \\
  & Jitter & 79.17 & 88.88 & 85.48 & 85.76 & 92.00 & 79.81 & 80.85 & 87.63 & 82.89 & 78.50 \\
  & Semantic Jitter~\cite{aron-iccv17} & 83.10 & 89.20 & 88.76 & \textbf{88.49} & \textbf{94.10} & 82.37 & 85.08 & \textbf{89.07} & 86.31 & 82.26 \\
  \Xcline{2-12}{0.3pt}
  & \Gape[3pt]{ATTIC (ours)} & 83.80 & \textbf{89.51} & \textbf{89.23} & 88.18 & \textbf{94.10} & \textbf{85.85} & \textbf{86.41} & 88.04 & \textbf{87.78} & 83.07 \\
  \Xhline{1pt}
  \multirow{5}{*}[0ex]{\rotatebox{90}{\textbf{Auto}}} & Real~\cite{Singh16} & 84.26 & 88.58 & 88.99 & 88.18 & 94.10 & 82.83 & 83.96 & 88.25 & 84.35 & 83.87 \\
  & Real+ & 81.71 & 87.96 & 87.12 & 87.58 & 91.05 & 82.60 & 84.41 & 87.63 & 85.82 & 83.87 \\
  & Jitter & 82.87 & 87.96 & 86.65 & 87.58 & 93.91 & 80.51 & 84.63 & 88.66 & 83.37 & 79.84 \\
  & Semantic Jitter~\cite{aron-iccv17} & 84.72 & 88.89 & 89.70 & 89.39 & 94.29 & 83.99 & 86.41 & 89.07 & 85.82 & 83.87 \\
  \Xcline{2-12}{0.3pt}
  & \Gape[3pt]{ATTIC (ours)} & \textbf{87.04} & \textbf{89.20} & \textbf{91.57} & \textbf{91.21} & \textbf{94.48} & \textbf{87.94} & \textbf{87.31} & \textbf{89.90} & \textbf{86.31} & \textbf{85.75} \\
  \Xhline{1pt}
  \end{tabular}
  \renewcommand{\arraystretch}{1}
  \end{center}
  \vspace*{-0.1in}
  \caption{Accuracy for the 10 attributes in the Shoes dataset.  ``Normal" means that the synthetic images generated by Semantic Jitter and our method are labeled by human annotators.  ``Auto" means that an additional $n$ \emph{unlabeled} synthetic image pairs are added for all methods except Real; those images adopt their inferred attribute labels.  In all cases, all methods use exactly $n$ total labels. The row for the Real baseline is repeated for Normal and Auto for easier comparison purposes.}
  \label{tab:onebatch_shoes}
  \vspace*{0.15in}
\end{table*}


\begin{table*}
  \begin{center}\footnotesize
  \renewcommand{\arraystretch}{1.2}
  \begin{tabular}{|c||l||*{8}{c}|}
  \cline{2-10}
  \multicolumn{1}{c|}{} & & \Gape[4pt]{\textbf{Bald}} & \textbf{DarkHair} & \textbf{BigEyes} & \textbf{Masculine} & \textbf{MouthOpen} & \textbf{Smiling} & \textbf{Forehead} & \textbf{Young} \\
  \Xhline{1pt}
  \multirow{5}{*}[0ex]{\rotatebox{90}{\textbf{Normal}}} & Real~\cite{Singh16} & 79.80 & 86.77 & 78.18 & 92.96 & 87.50 & 74.44 & 80.00 & 78.76 \\
  & Real+ & 81.82 & 86.03 & 80.00 & 92.96 & 86.67 & 75.94 & 81.21 & 79.28 \\
  & Jitter & 80.81 & 85.29 & 76.36 & 88.73 & 77.50 & 74.44 & 81.05 & 77.20 \\
  & Semantic Jitter~\cite{aron-iccv17} & 81.82 & 87.50 & 83.64 & 92.96 & \textbf{88.33} & 79.70 & 83.16 & \textbf{81.35}  \\
  \Xcline{2-10}{0.3pt}
  & \Gape[3pt]{ATTIC (ours)} & \textbf{84.85} & \textbf{88.24} & \textbf{85.46} & \textbf{95.78} & 79.17 & \textbf{81.96} & \textbf{84.21} & 80.31 \\
  \Xhline{1pt}
  \multirow{5}{*}[0ex]{\rotatebox{90}{\textbf{Auto}}} & Real~\cite{Singh16} & 79.80 & 86.77 & 78.18 & 92.96 & \textbf{87.50} & 74.44 & 80.00 & 78.76 \\
  & Real+ & 81.82 & 86.03 & 80.00 & 92.96 & 86.67 & 75.94 & 81.21 & 79.28 \\
  & Jitter & 81.82 & 86.87 & 80.00 & 91.55 & 84.17 & 74.44 & \textbf{85.26} & \textbf{79.79} \\
  & Semantic Jitter~\cite{aron-iccv17} & 82.83 &	88.24 & 81.82 & \textbf{94.37} & 85.00 & 75.19 & 83.16 & 79.28 \\
  \Xcline{2-10}{0.3pt}
  & \Gape[3pt]{ATTIC (ours)} & \textbf{85.86} & \textbf{90.44} & \textbf{83.36} & \textbf{94.37} & 82.50 & \textbf{76.69} & \textbf{85.26} & 78.24 \\
  \Xhline{1pt}
  \end{tabular}
  \renewcommand{\arraystretch}{1}
  \end{center}
  \vspace*{-0.1in}
  \caption{Accuracy for the 8 attributes in the Faces dataset.  Format is the same as Table~\ref{tab:onebatch_shoes}.}
  \label{tab:onebatch_faces}
\end{table*}

\vspace*{0.12in}
\section{Experiments}
\label{sec:experiments}

To validate our active generation approach, we explore two fine-grained attribute domains: shoes and faces.  To our knowledge, these are the only existing datasets with sufficient \emph{instance-level} relative attribute labels.

\subsection{Experimental Setup}

\vspace*{0.05in}
\noindent
\textbf{Datasets:}  We use two publicly available datasets, UT-Zappos50K~\cite{aron-cvpr14} and LFW~\cite{lfw}.  For each, our method uses real images to initialize training and then creates its own synthetic training images.  Please note that the synthetic images are labeled by human annotators before they are used to augment the training set.

\begin{itemize}
  \vspace*{0.08in}
  \item  \textbf{Catalog Shoe Images:}  We use the \textbf{UT-Zappos50K} dataset~\cite{aron-cvpr14} with the fine-grained attributes from~\cite{aron-iccv17}.  The dataset consists of over 50,000 shoe images from Zappos.com.  There are 10 attributes (\textit{comfort}, \textit{casual}, \textit{simple}, \textit{sporty}, \textit{colorful}, \textit{durable}, \textit{supportive}, \textit{bold}, \textit{sleek}, and \textit{open}), each with about 4,000 labeled pairs.
  \vspace*{0.08in}
  \item  \textbf{Human Face Images:}  We use the \textbf{LFW} dataset~\cite{lfw} and the \textbf{LFW-10} dataset~\cite{Sandeep14}.  LFW consists of over 13,000 celebrity face images from the wild along with real-valued labels on 73 attributes.  LFW-10 consists of a subset of 2,000 images from LFW along with relative labels on 10 attributes.  We use the 8 attributes (\textit{bald}, \textit{dark hair}, \textit{big eyes}, \textit{masculine}, \textit{mouth open}, \textit{smiling}, \textit{visible forehead}, and \textit{young}) in the intersection of these two datasets.  For the real image pairs, there are about 500 labeled pairs per attribute from LFW-10.
  \vspace*{0.08in}
\end{itemize}

\noindent
All images for all methods are resized to $64\times64$ pixels to match the output resolution of the image generator.  For all experiments, we only use high quality (high agreement/high confidence) relative labels, following the protocol of~\cite{aron-iccv17}.  We collect annotations for our method's automatically generated training pairs using Mechanical Turk; we obtain 5 worker responses per label and take the majority vote.  Workers are free to vote for discarding a pair if they find it illegible, which happened for just 17\% of the generated pairs.  See Section~\ref{sec:appendix} for the data collection interface and instructions to annotators.

\vspace*{0.1in}
\noindent
\textbf{Implementation Details:}  During training, we validate all hyperparameters (such as the learning rate, the learning rate decay, and the weight decay) on a separate validation set.  We run all experiments (including individual batches) to convergence or to a max of 250 and 100 epochs for shoes and faces, respectively.

For the individual modules, implementation details are as follows.  Ranker: We pre-train the DeepSTN ranking network without the global image channel using only the real image pairs (see~\cite{Singh16} for details on the two rounds of training).  Generator:  We use the code provided by the authors of Attribute2Image~\cite{Yan16-eccv} with all default parameters.  We pre-train the image generators using a disjoint set of real images (38,000 and 11,000 images for shoes and faces, respectively) that do not have any associated relative labels.  We use the trained decoder in ATTIC while keeping the parameters constant throughout end-to-end training for the ranker and control (i.e., learning rate of zero on decoder).  Control:  We initialize the layers using ReLU initialization~\cite{He15}.  The learning rate decays such that as learning goes on, the changes to $\bm{y},\bm{z}$ become smaller.


\vspace*{0.15in}
\noindent
\textbf{Baselines:}  We consider the following baselines.
\begin{itemize}
  \vspace*{0.05in}
  \item \textbf{Semantic Jitter}~\cite{aron-iccv17}, an existing approach for data augmentation that uses synthetic images altered by one attribute at a time using a manually defined sampling policy.  We use the authors' provided synthetic shoe image pairs, Zap50K-Synth, for this purpose.  It contains 2,000 labeled pairs per attribute.  For the synthetic face image pairs, we collect relative labels on 1,000 pairs per attribute.
  \vspace*{0.05in}
  \item \textbf{Jitter}, the traditional data augmentation process where the real images are jittered through low-level geometric and photometric transformations.  We follow the jitter protocol defined in~\cite{Dosovitskiy14}, which includes translation, scaling, rotation, contrast, and color.  The jittered image pairs retain the corresponding real pairs' respective labels.
  \vspace*{0.05in}
  \item  \textbf{Real}, standard approach which trains with only real labeled image pairs.
  \vspace*{0.05in}
  \item  \textbf{Real+}, slight modification that adds real image pairs with their pseudo labels to Real.  We form pseudo pairs of equal size bootstrapped from the attribute strength of the images that are used to train the image generator.  The purpose of this baseline is to ensure that our advantage is not due to our network's access to the attribute-strength labeled images that the image generator module requires for training.
  \vspace*{0.08in}
\end{itemize}

\noindent
Note that all methods use the same state-of-the-art ranking network for training and predictions, hence any differences in results will be attributable to the training data and augmentation strategy.

\vspace*{0.1in}
\subsection{Competing with Real Training Images}

First we test our hypothesis that restricting models to only available Web photos for training can be limiting.  We compare the data augmentation baselines and ATTIC to the Real baselines, where all methods are given the exact same amount of total manual annotations.  The Real, Real+, and Jitter baselines use all $n$ available real labeled image pairs.  Semantic Jitter and ATTIC use \emph{half} of the real labeled image pairs ($\frac{n}{2}$), then augment those pairs with $\frac{n}{2}$ manually labeled synthetic image pairs that they generate.

Tables~\ref{tab:onebatch_shoes} and~\ref{tab:onebatch_faces} (Normal) show the results for the Shoes and Faces datasets, respectively.  Though using exactly the same amount of manual labels as the Real baseline, our method nearly always outperforms it.  This shows that simply having more real image pairs labeled is not always enough; our generated samples improve the training across the variety of attributes in ways the existing real image pairs could not.  In addition, we see from Real+ that the image generator training images have only a marginal (and sometimes negative) effect on the baseline's results.  This indicates that both Real and Real+ suffer from the same sparsity issue, as the images are taken from similar pool of real images.  The addition of similarly distributed (real) images lacks the fine-grained details needed to train a stronger model.  Furthermore, our approach also outperforms (or matches) Semantic Jitter in 8 out of 10 shoe attributes and 6 out of 8 face attributes, with gains of just over 3\% in some cases.  This demonstrates our key advantage over Semantic Jitter~\cite{aron-iccv17}, which is to actively adapt the generated images to best suit the learning of the model, as opposed to what looks the best to human eyes.  Unlike~\cite{aron-iccv17}, which manually modifies one attribute at a time, our approach can modify multiple attributes simultaneously in a dynamic manner, accounting for their dependencies.

In Tables~\ref{tab:onebatch_shoes} and~\ref{tab:onebatch_faces} we also consider an ``Auto'' scenario where instead of adding the $\frac{n}{2}$ generated images with their manual annotations, we bootstrap from all $n$ real labeled image pairs. Then, we generate another $n$ synthetic images and---rather than get them labeled---simply adopt their inferred attribute comparison labels.  In this case, the ``ground truth'' ordering for attribute $j$ for generated images $\hat{\bm{x}}^A$ and $\hat{\bm{x}}^B$ is automatically determined by the magnitudes of their associated parameter values $\bm{y}^A(j)$ and $\bm{y}^B(j)$ output by the control module.  As before, Jitter adopts the label of the source pair it jittered.  Once again, all methods use the same number of labels.

Table~\ref{tab:onebatch_shoes} and~\ref{tab:onebatch_faces} (Auto) show the results.  Our model performs even a bit better in this setting, suggesting that the inferred labels are often accurate, and the extra volume of ``free" training pairs is helpful.  We outperform (or match) Semantic Jitter in all 10 shoe attributes and 6 out of 8 face attributes.  Jitter suffers a slight performance boost sometimes, but can even be detrimental on these datasets.

While our method performs well overall, for a couple of attributes (i.e \emph{mouth-open}, \emph{young}) we underperform both Real and Semantic Jitter.  Upon inspection, we find our weaker performance there is due to deficiency in the image generators.  In such scenarios, Semantic Jitter may have an advantage with its manual selection process.


\begin{figure}
  \centering
  \begin{subfigure}[t]{.4\textwidth}
    \includegraphics[width=\textwidth]{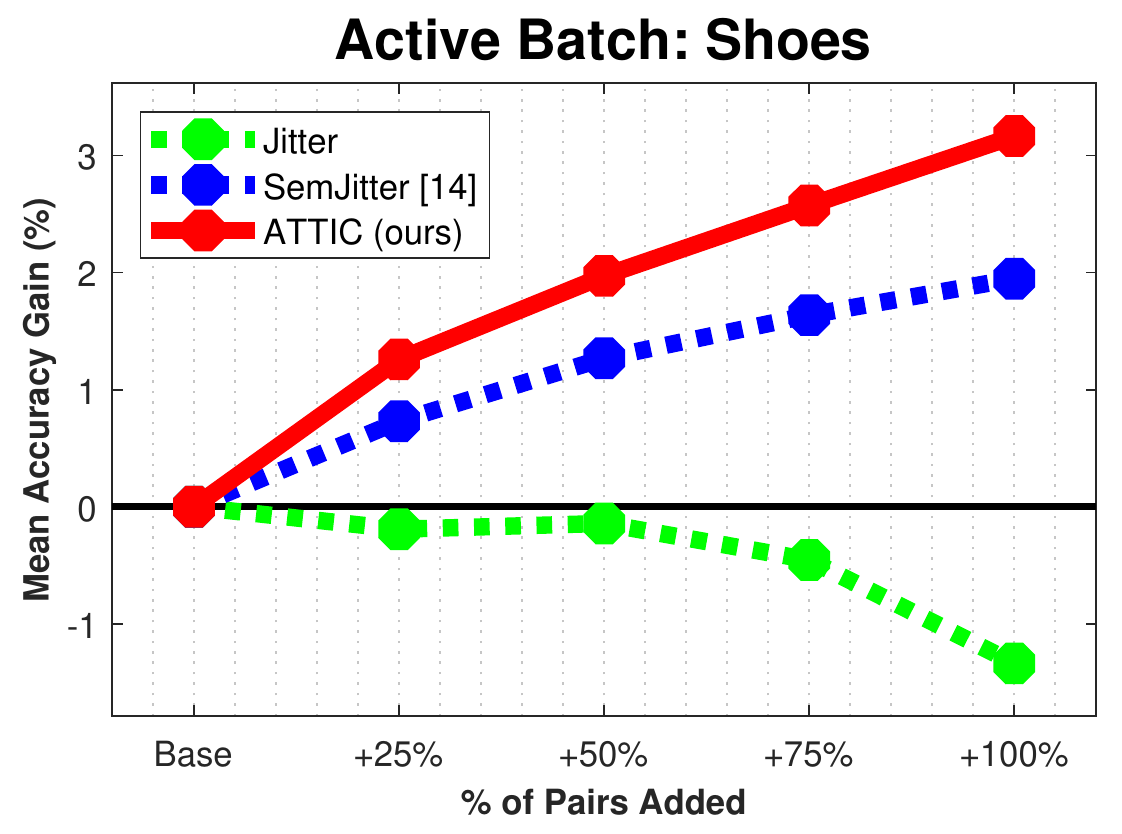}
  \end{subfigure}\\
  \begin{subfigure}[t]{.4\textwidth}
    \includegraphics[width=\textwidth]{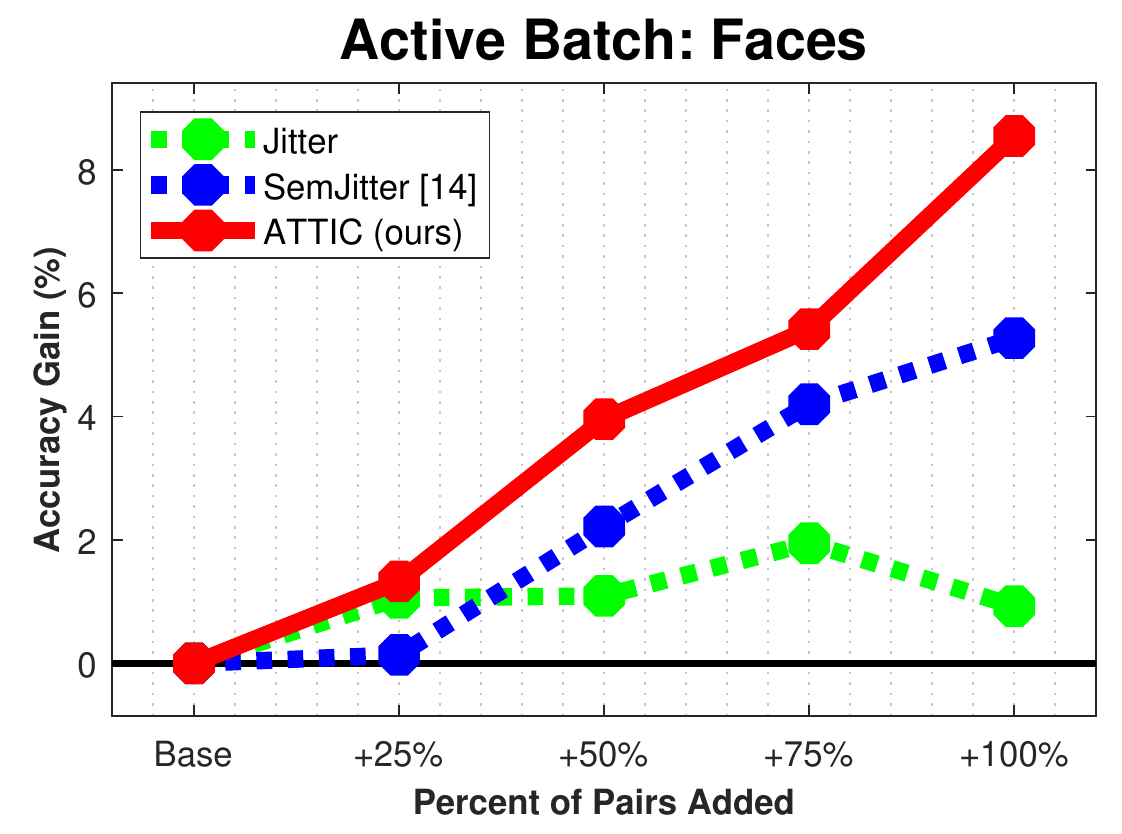}
  \end{subfigure}
  \caption{Active learning curves for the Shoes (top) and Faces (bottom) datasets.  We show the average gain over the Real baseline after each batch of additional generated image pairs.  Our approach nearly doubles the gain achieved by Semantic Jitter for both domains.}
  \vspace*{0.1in}
  \label{fig:abatch_gain}
\end{figure}

\subsection{Active vs.~Passive Training Image Generation}

Next we examine more closely ATTIC's active learning behavior.  In this scenario, we suppose the methods have exhausted all available real training data (i.e., we use all $n$ real labeled image pairs to initialize the model), and our goal is to augment this set.  We generate the synthetic (labeled) image pairs in batches (again, not to be confused with the mini-batches when training neural networks).  After each batch, we have them annotated, update the ranker's training set, and re-evaluate it on the test set.  The weights of the control module are carried over from batch to batch, while the ranker module restarts at its pre-trained state at the beginning of each batch.

Figure~\ref{fig:abatch_gain} shows the results for both datasets.  We plot active learning curves to show the accuracy improvements as a function of annotator effort---steeper curves are better, as they mean the system gets more accurate with less manual labeling.  We see the average gains of our approach over the Real baseline increase most sharply compared to the baselines.  Our approach achieves a gain of over 3\% and 8\% for the two domains, respectively, which is almost double that of the Semantic Jitter baseline.  Jitter falls short once again, suggesting that traditional low-level jittering has limited impact in these fine-grained ranking tasks.  Please see Section~\ref{sec:appendix} for the individual performance plots of each attribute.

\begin{table}[t]
  \begin{center}\footnotesize
  \renewcommand{\arraystretch}{1.2}
  \begin{tabular}{|c||l||c c c|}
  \cline{2-5}
  \multicolumn{1}{c|}{} & & \Gape[4pt]{\textbf{Open}} & \textbf{Sporty} & \textbf{Comfort} \\
  \Xhline{1pt}
  \multirow{5}{*}{\rotatebox[origin=c]{90}{\textbf{Zap50K-1}}} & RelAttr~\cite{relattr11} & 88.33 & 89.33 & 91.33 \\
  & FG-LP~\cite{aron-cvpr14} & 90.67 & 91.33 & 93.67 \\
  & DeepSTN~\cite{Singh16} & 93.00 & 93.67 & 94.33 \\
  & SemJitter~\cite{aron-iccv17} & 95.00 & \textbf{96.33} & 95.00 \\
  \cline{2-5}
  & \Gape[3pt]{ATTIC (ours)} & \textbf{95.67} & 96.00 & \textbf{95.67} \\
  \Xhline{1pt}
  \multirow{5}{*}{\rotatebox[origin=c]{90}{\textbf{Zap50K-2}}} & RelAttr~\cite{relattr11} & 60.36 & 65.65 & 62.82 \\
  & FG-LP~\cite{aron-cvpr14} & 69.36 & 66.39 & 63.84 \\
  & DeepSTN~\cite{Singh16} & 70.73 & 67.49 & 66.09 \\
  & SemJitter~\cite{aron-iccv17} & \textbf{72.18} & 68.70 & 67.72 \\
  \cline{2-5}
  & \Gape[3pt]{ATTIC (ours)} & 71.68 & \textbf{69.62} & \textbf{68.64} \\
  \hline
  \end{tabular}
  \renewcommand{\arraystretch}{1}
  \end{center}
  \vspace*{-0.1in}
  \caption{Extension to the result table from~\cite{aron-iccv17} that includes our results for the same Zappos50K splits.  All methods are trained and tested on $64 \times 64$ images for an apples-to-apples comparison.}
  \label{tab:zap50k_rerun}
  \vspace*{0.1in}
\end{table}

\vspace*{-0.05in}
\subsection{Comparison to Previously Published Results}

The experiments thus far demonstrate that our approach allows more accurate fine-grained predictions for the same amount of manual annotation effort, compared to both traditional training procedures with real images as well as existing jitter approaches.  Next we present results for our approach alongside all available comparable reported results on the Shoes data.

Table~\ref{tab:zap50k_rerun} shows the results reported in~\cite{aron-iccv17} alongside our results using the same UT-Zappos50K train/test split.  Note that there are only three relevant attributes here (\emph{open}, \emph{sporty}, \emph{comfort}), as opposed to the 10 we report above for Shoes, because prior work includes only these three due to data split restrictions. Following~\cite{aron-iccv17}, for an apples-to-apples comparison, all methods are applied to the same 64 $\times$ 64 images.  Our results use the method exactly as described above for the ``Auto'' scenario.  

Our method outperforms all the existing methods for the majority of the attributes.  Semantic Jitter~\cite{aron-iccv17} outperforms ours for \emph{sporty} in the first test set and \emph{open} in the second test set, indicating that those attributes were similarly well-served by that method's heuristic choice for generated images.  However, our automated method overall has the advantage.

\vspace*{-0.05in}
\subsection{Qualitative Analysis}

As we have seen in the results above, the synthetic image pairs generated by our approach outperform those selected by the heuristic and passive selection processes of Semantic Jitter and Jitter in almost all scenarios.
The advantage of our active generation approach is its ability to modify the generated image pairs in a way that is best for the learning of the model.

\begin{figure}[t]
  \centering
  \begin{subfigure}[t]{.49\textwidth}
    \includegraphics[width=\textwidth]{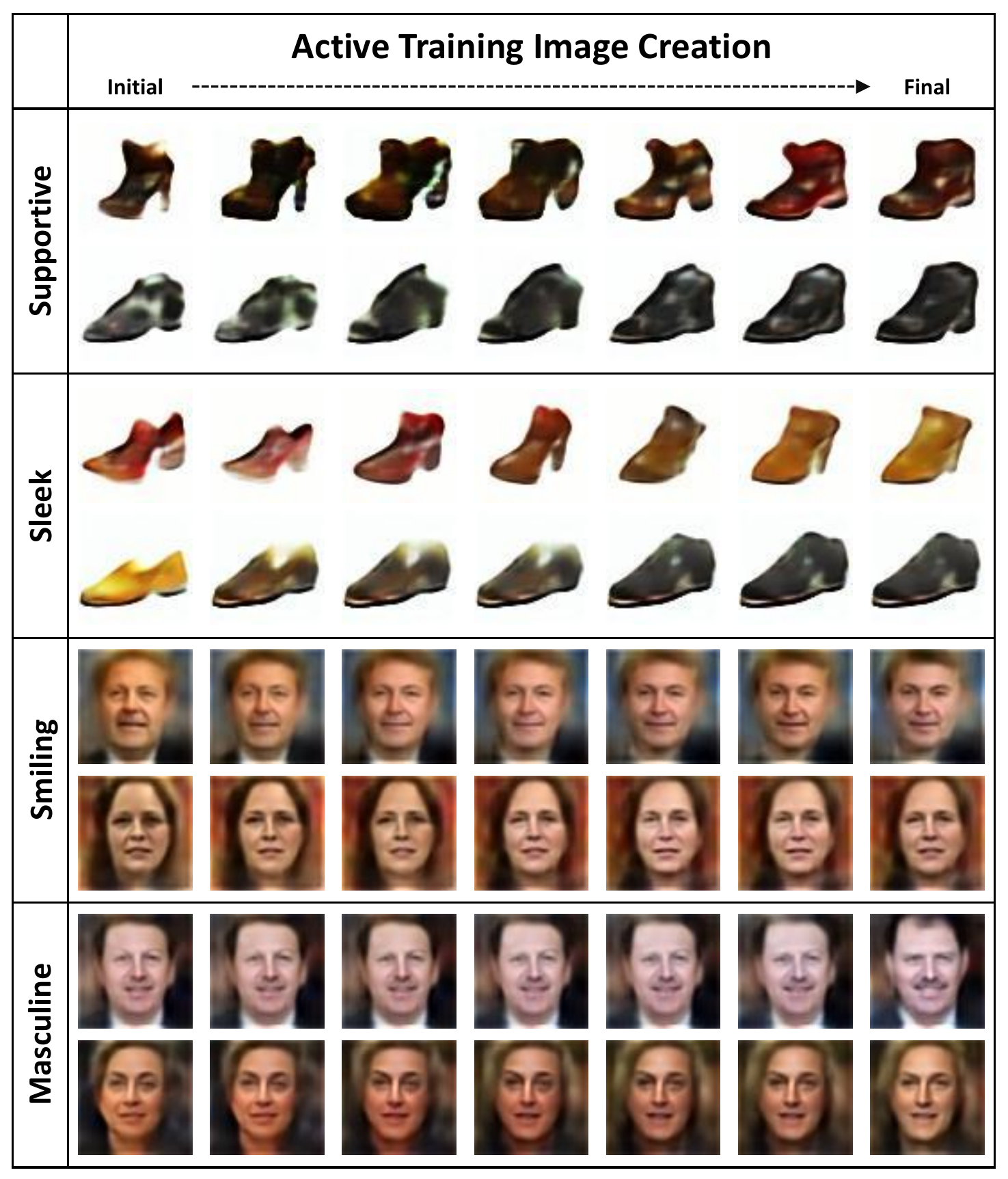}
  \end{subfigure}
  \vspace*{-0.2in}
  \caption{Visualization of the progression of some synthetic image pairs $(\hat{\bm{x}}^A, \hat{\bm{x}}^B)$ during training.  Our model learns patterns between all the attributes, modifying multiple attributes simultaneously. For example, while modifying the face images for the attribute \emph{masculine} (last row), our model learned to change the attribute \emph{smiling} as well.}
  \label{fig:spectrum}
  \vspace*{0.1in}
\end{figure}

Figure~\ref{fig:spectrum} shows examples of how the synthetic images look between the first and the last epoch of the training.  We can see that pairs generated by our approach demonstrate change in multiple attributes while still keeping the target attribute of comparison at the forefront.  Furthermore, the final pairs selected for labeling also demonstrate subtler visual differences than the initial pairs, suggesting that our model has indeed learned to generate ``harder'' pairs.

\begin{figure*}[t]
    \centering
    \includegraphics[width=0.98\textwidth]{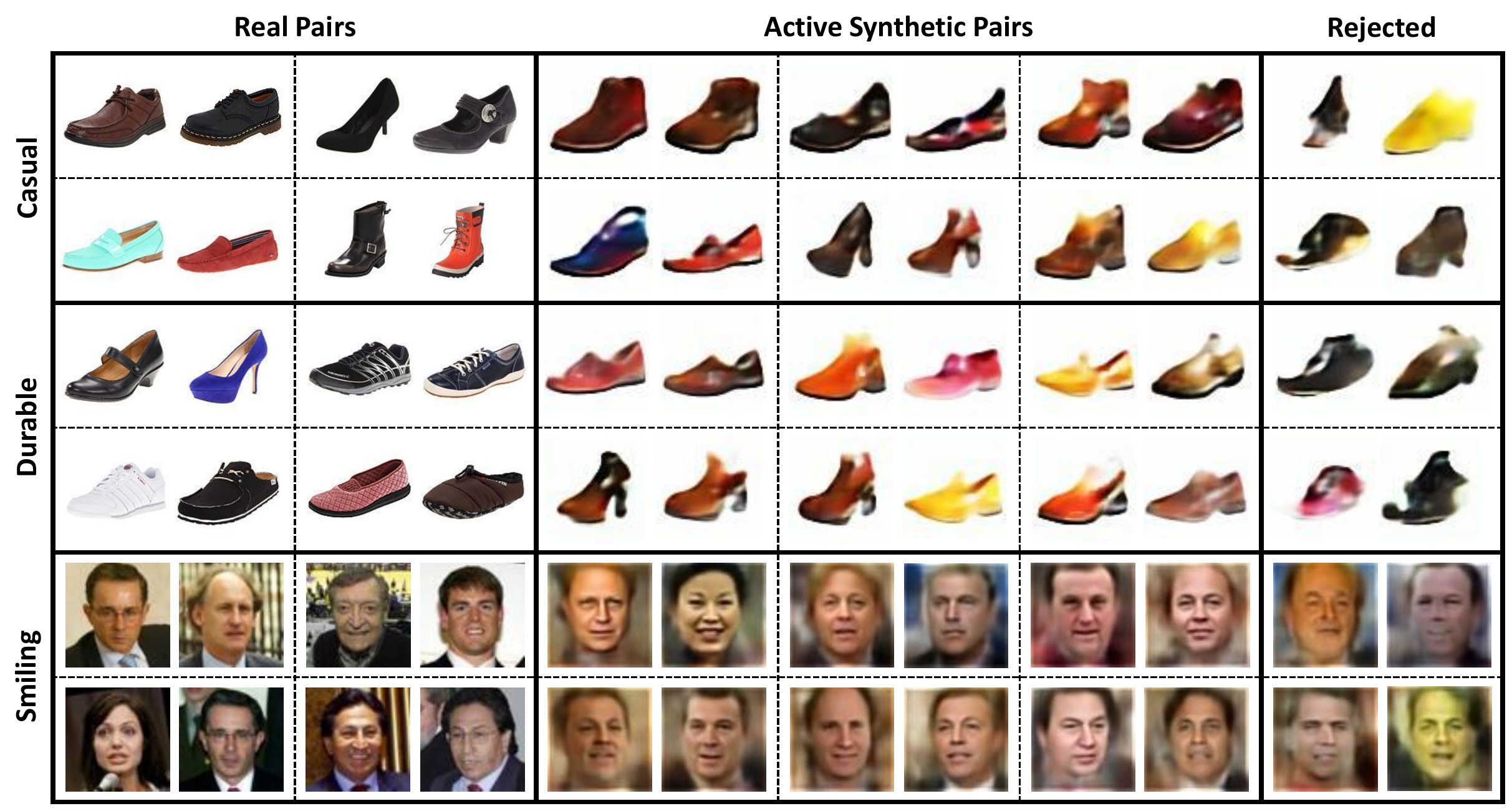}
    \vspace*{-0.05in}
    \caption{Sample training image pairs generated by ATTIC.  ``Harder'' real pairs that are incorrectly predicted by the baseline model (left).  Synthetic image pairs generated by our active approach (middle).  Synthetic image pairs that are rejected by the human annotators during the labeling process as illegible (right).}
    \vspace*{0.05in}
    \label{fig:qualitative}
\end{figure*}

Figure~\ref{fig:qualitative} compares these ``harder'' pairs to those from the real image pairs.  Overall we see that the actively generated synthetic pairs tend to have fine-grained differences and/or offer visual diversity from the real training samples.  The righthand side of Figure~\ref{fig:qualitative} shows examples of generated pairs rejected by annotators as illegible, which occurs 17\% of the time.  The relative low rate of rejection is an encouraging sign for making active query synthesis viable for image labeling tasks.


\vspace*{0.1in}
\section{Conclusion}
\label{sec:conclusion}

We introduced an approach for actively generating training image pairs that can benefit a fine-grained attribute ranker.  Our approach lets the system think ``outside of the pool" in annotation collection, imagining its own training samples and querying human annotators for their labels.  On two difficult datasets we showed that the approach offers real payoff in accuracy for distinguishing subtle attribute differences, with consistent improvements over existing data augmentation techniques.  In future work we plan to explore joint multi-attribute models for the ranker and consider how human-in-the-loop systems like ours might allow simultaneous refinement of the image generation scheme.

\vspace*{0.1in}
\section*{Acknowledgements} 

We thank Qiang Liu, Bo Xiong, and Xinchen Yan for helpful discussions.

\vspace*{0.1in}
{\small
\bibliographystyle{IEEEtran}
\bibliography{strings,refs}
}


\clearpage

\begin{figure*}[h]
\centering
    \begin{subfigure}[t]{.19\textwidth}
        \includegraphics[width=\textwidth]{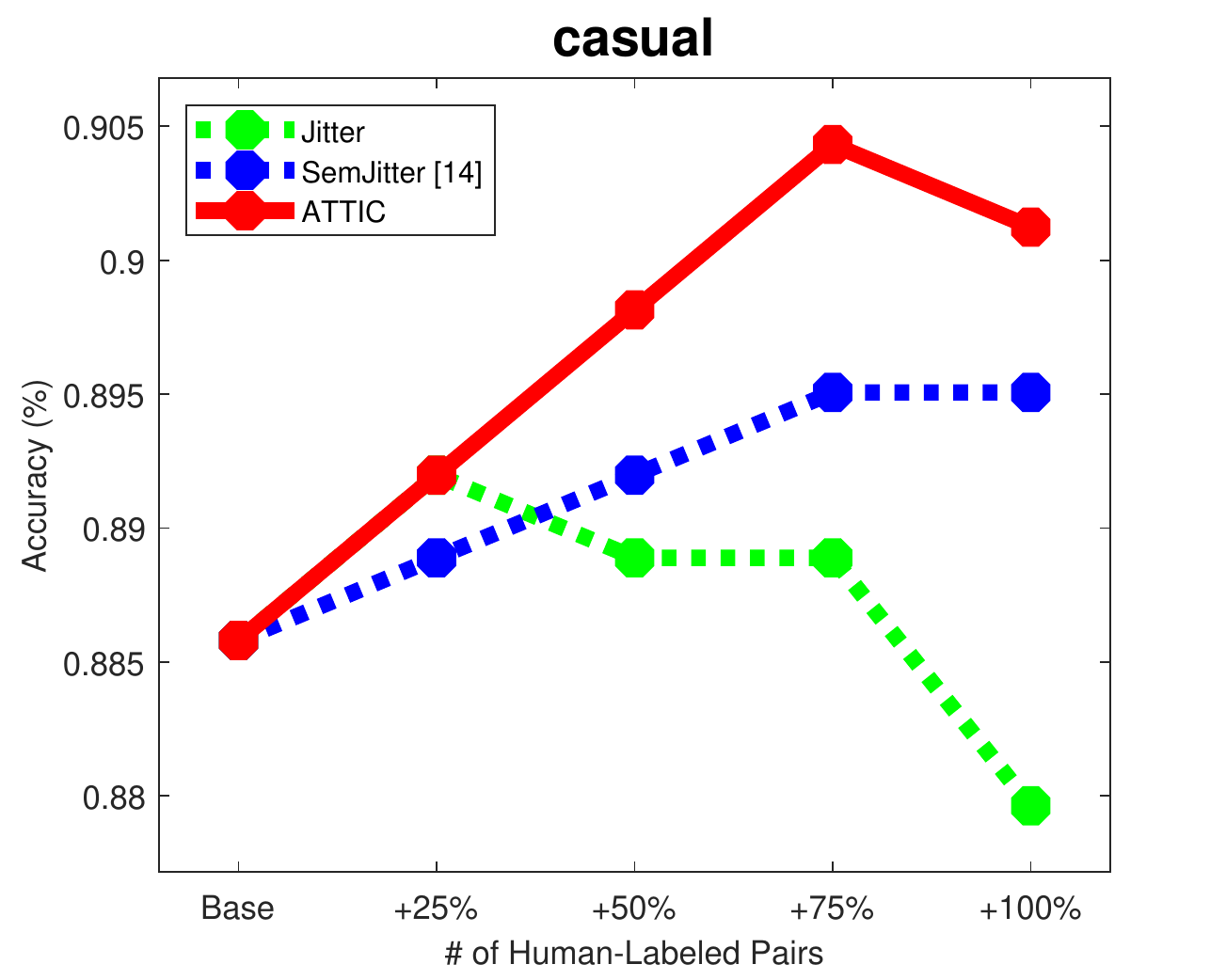}
    \end{subfigure}%
    \begin{subfigure}[t]{.19\textwidth}
        \includegraphics[width=\textwidth]{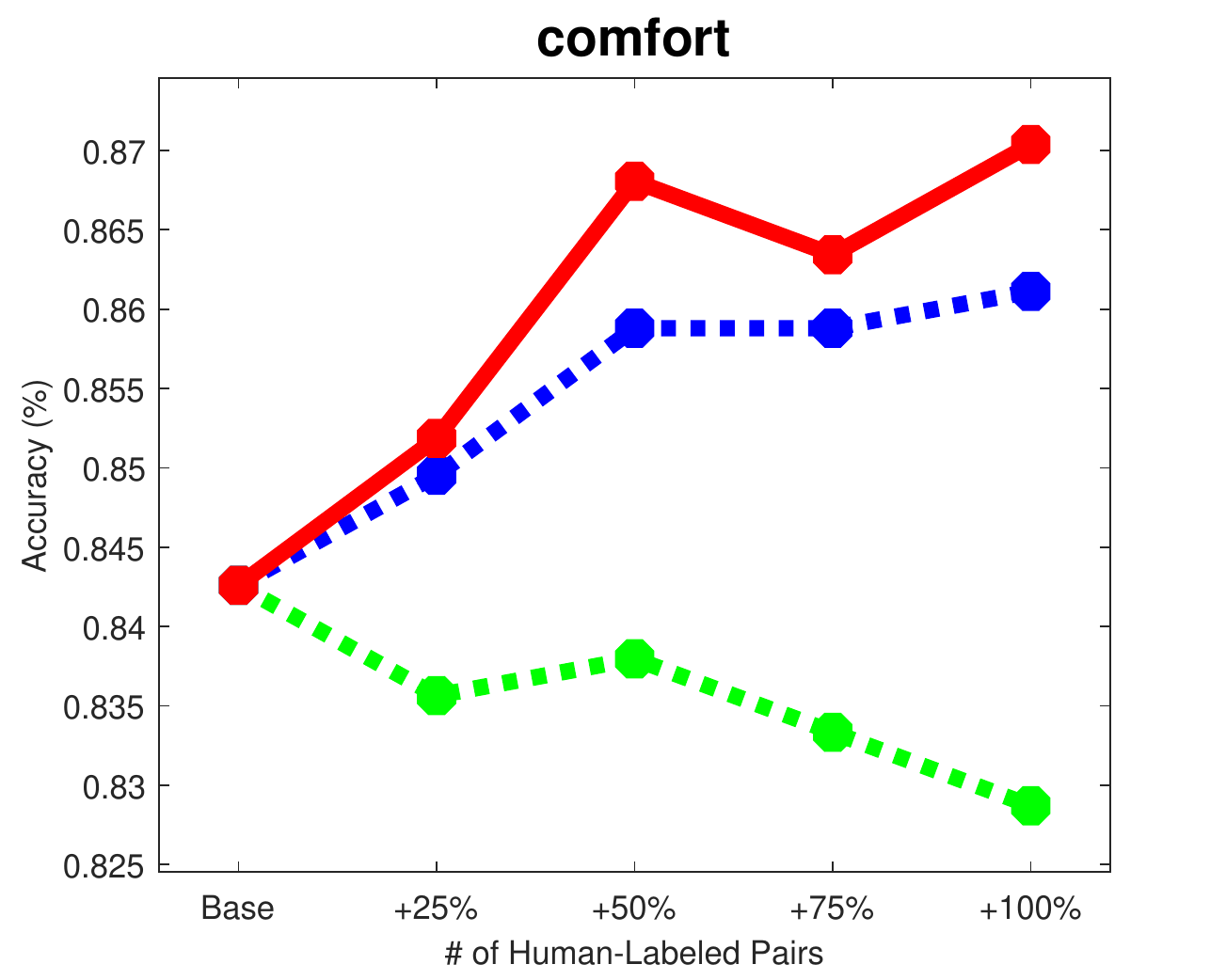}
    \end{subfigure}%
    \begin{subfigure}[t]{.19\textwidth}
        \includegraphics[width=\textwidth]{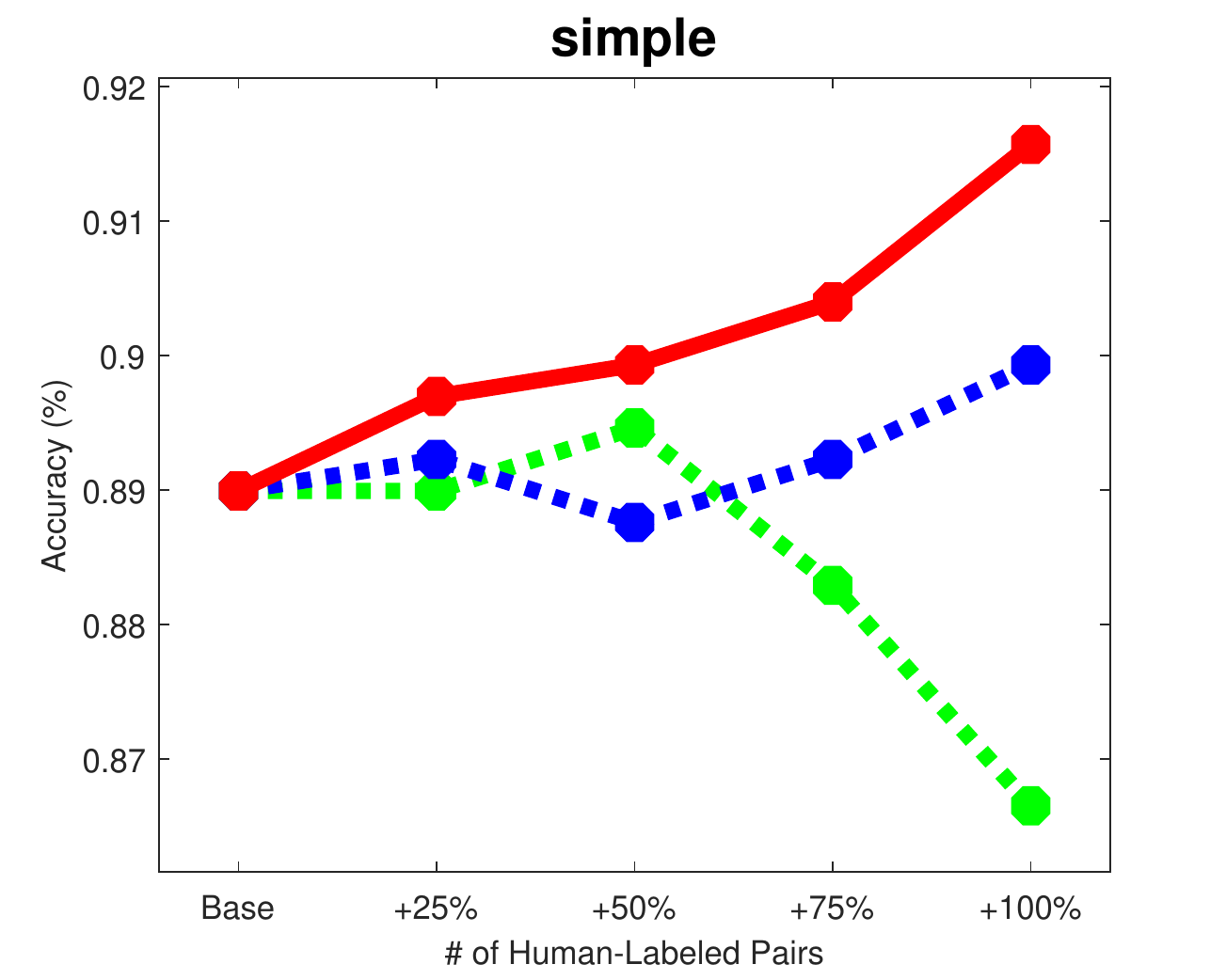}
    \end{subfigure}%
    \begin{subfigure}[t]{.19\textwidth}
        \includegraphics[width=\textwidth]{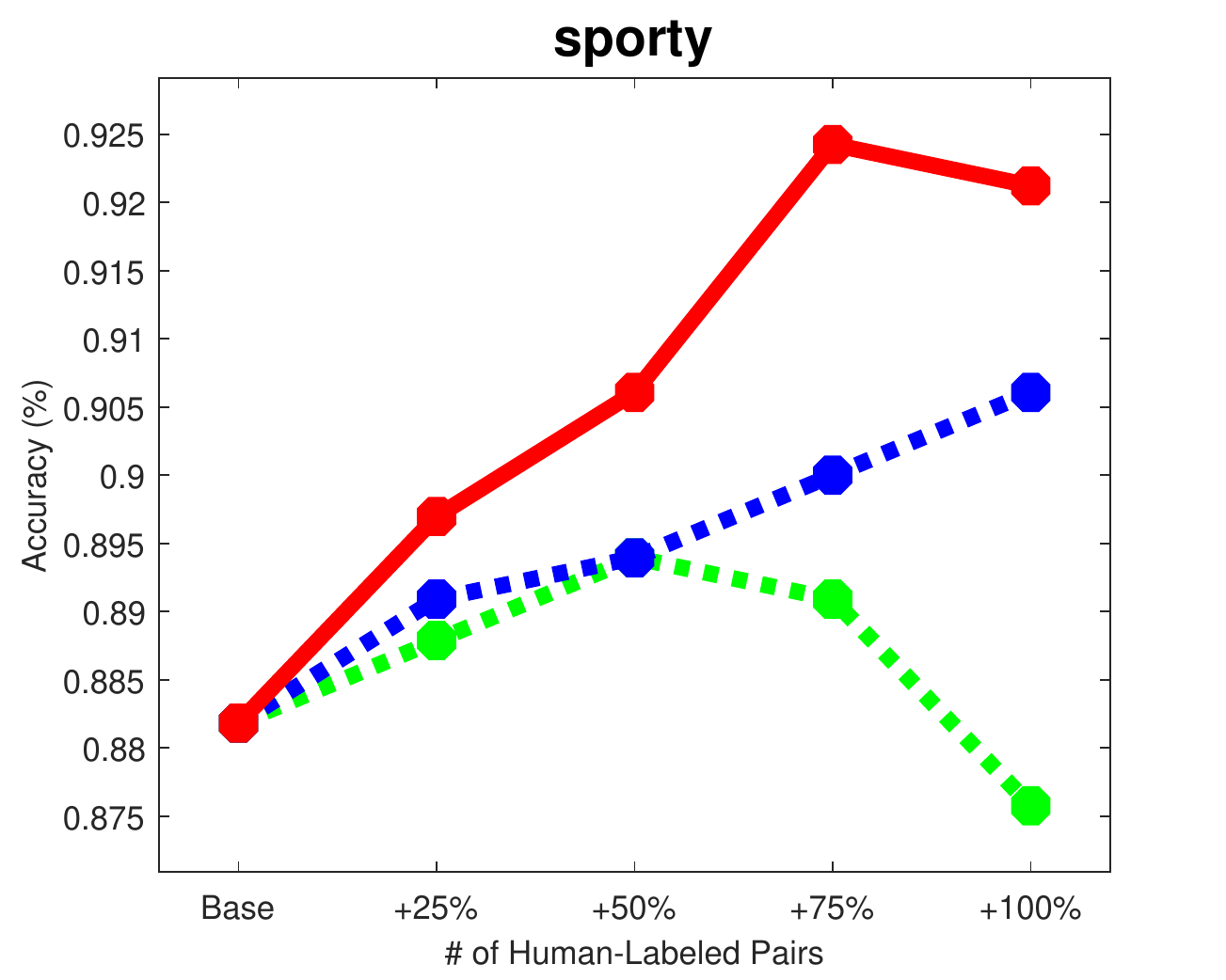}
    \end{subfigure}%
    \begin{subfigure}[t]{.19\textwidth}
        \includegraphics[width=\textwidth]{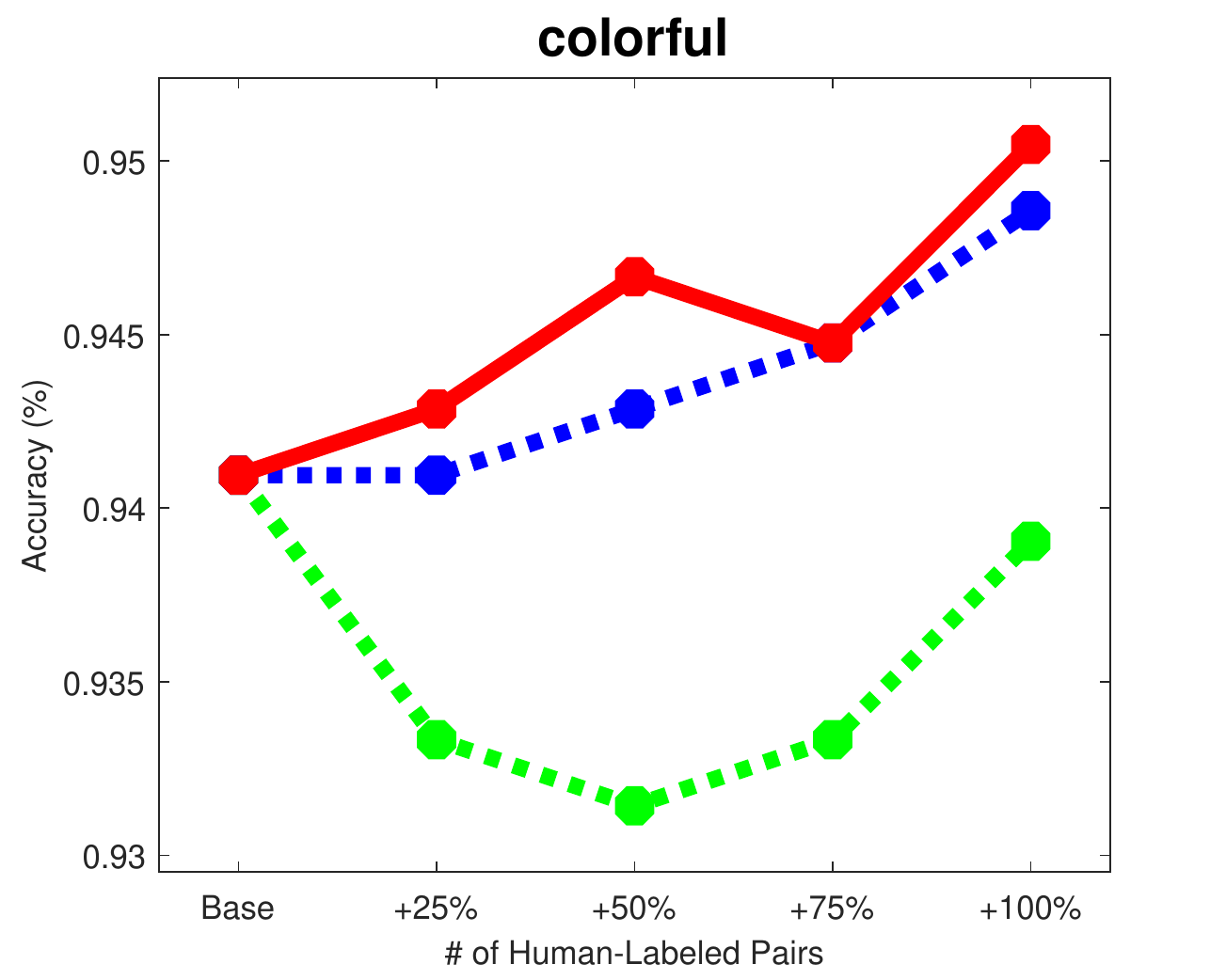}
    \end{subfigure}
    \begin{subfigure}[t]{.19\textwidth}
        \includegraphics[width=\textwidth]{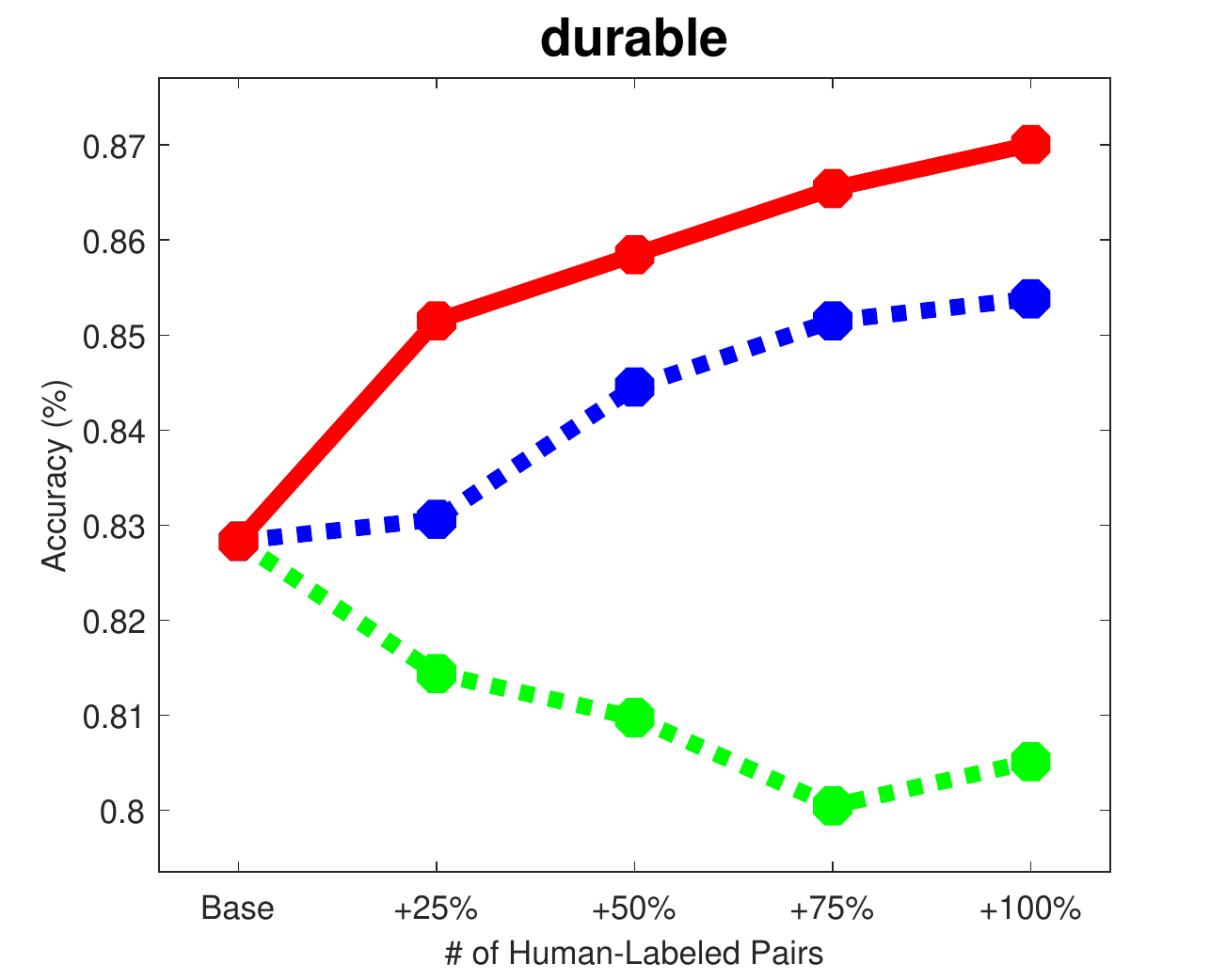}
    \end{subfigure}%
    \begin{subfigure}[t]{.19\textwidth}
        \includegraphics[width=\textwidth]{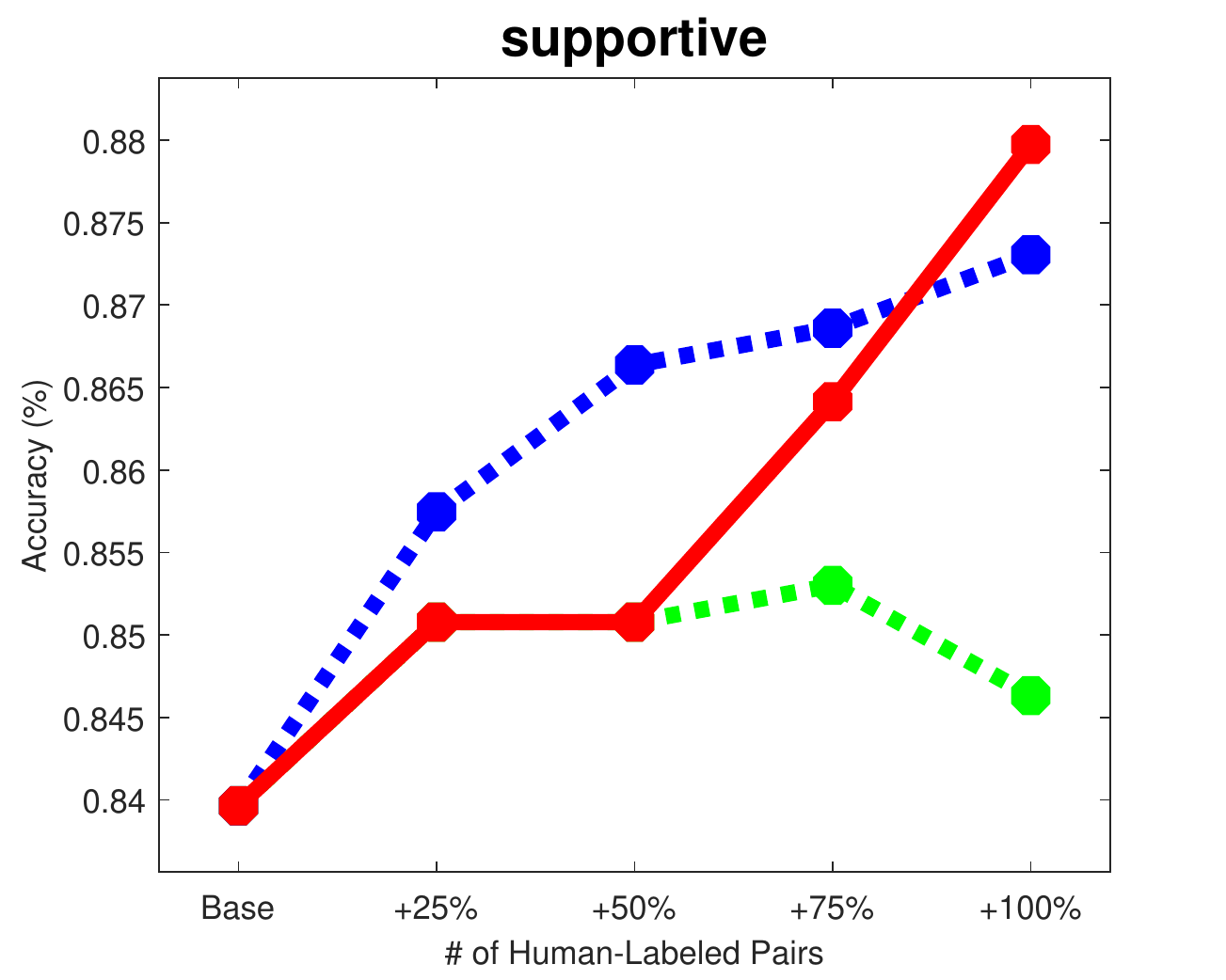}
    \end{subfigure}%
    \begin{subfigure}[t]{.19\textwidth}
        \includegraphics[width=\textwidth]{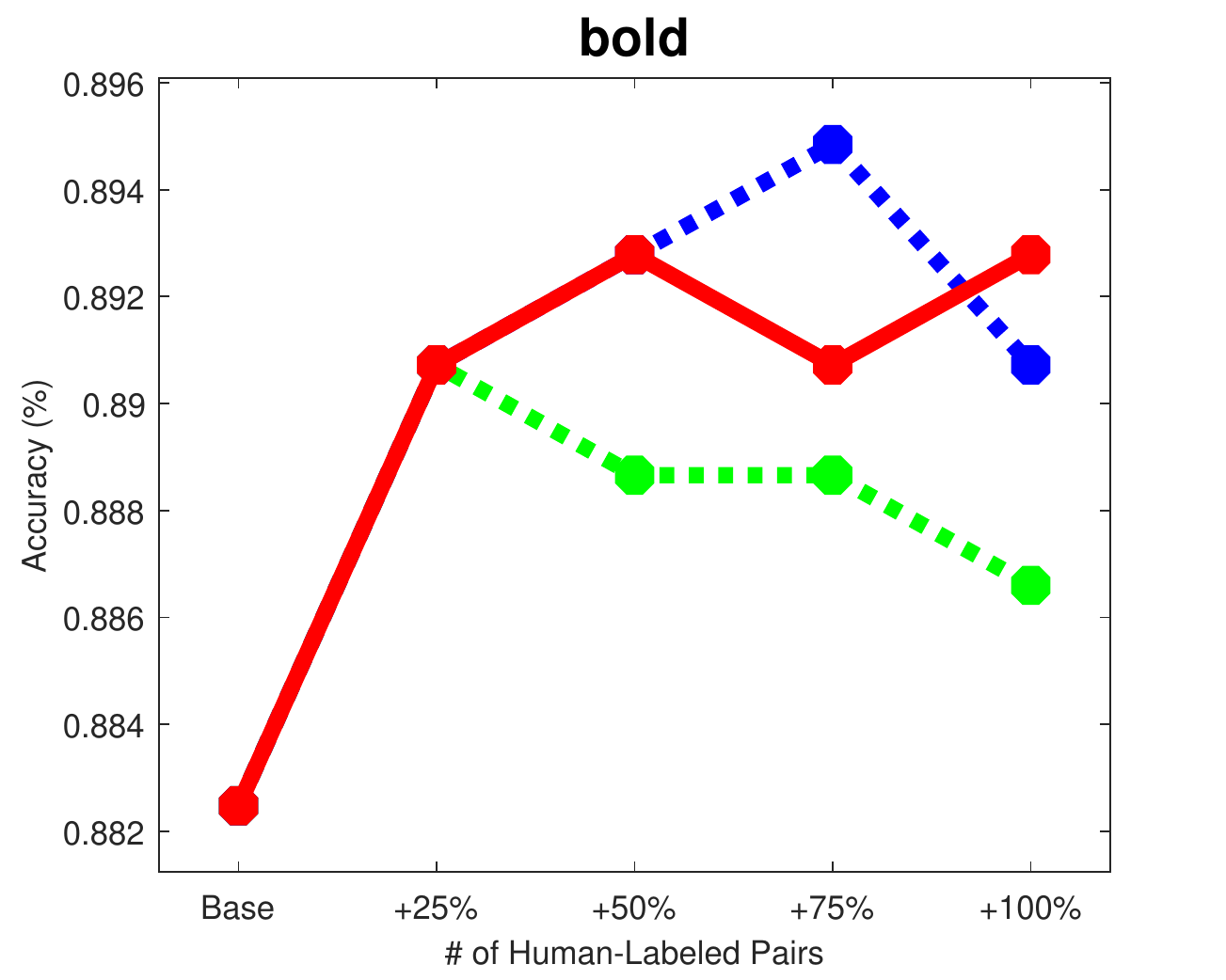}
    \end{subfigure}%
    \begin{subfigure}[t]{.19\textwidth}
        \includegraphics[width=\textwidth]{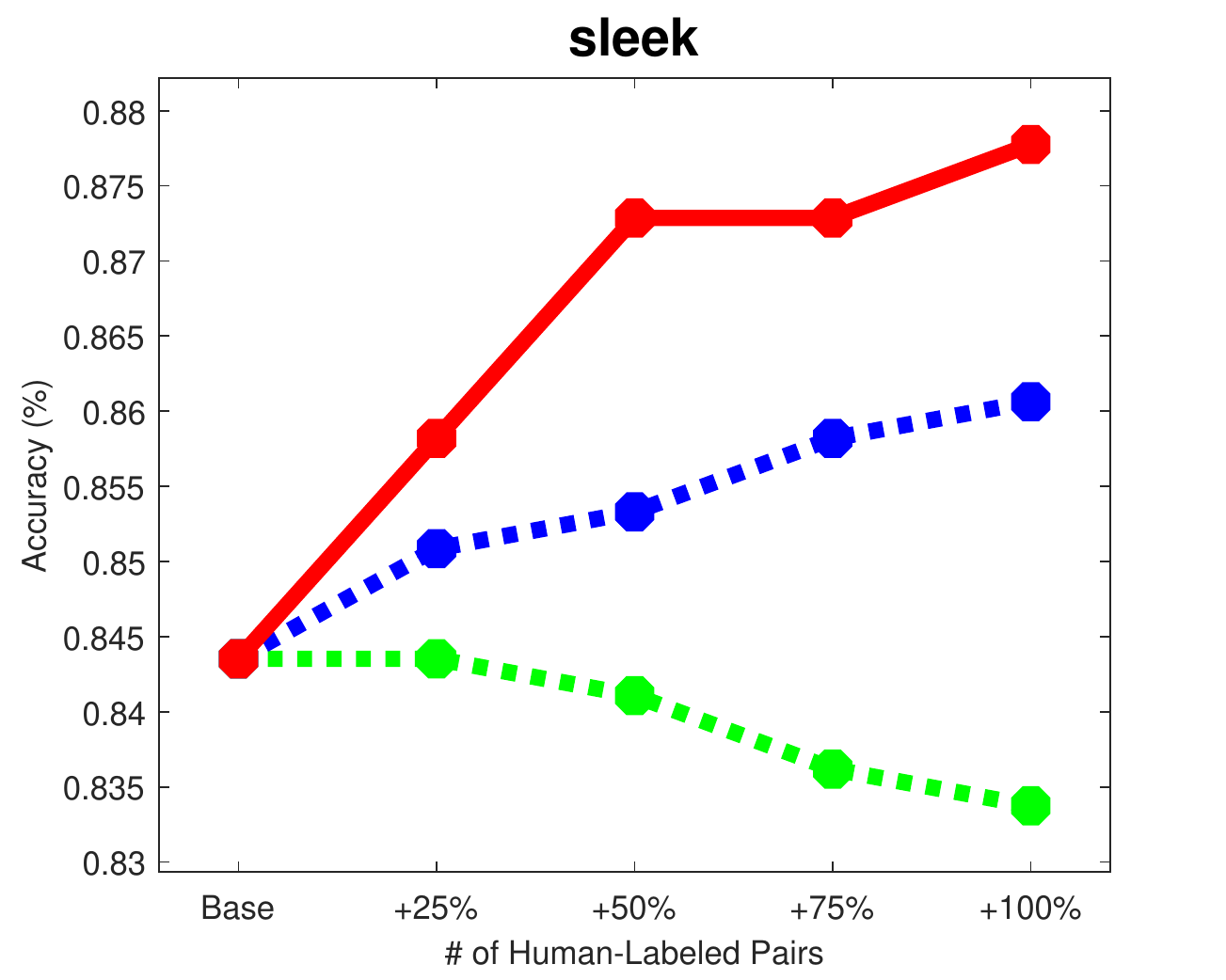}
    \end{subfigure}%
    \begin{subfigure}[t]{.19\textwidth}
        \includegraphics[width=\textwidth]{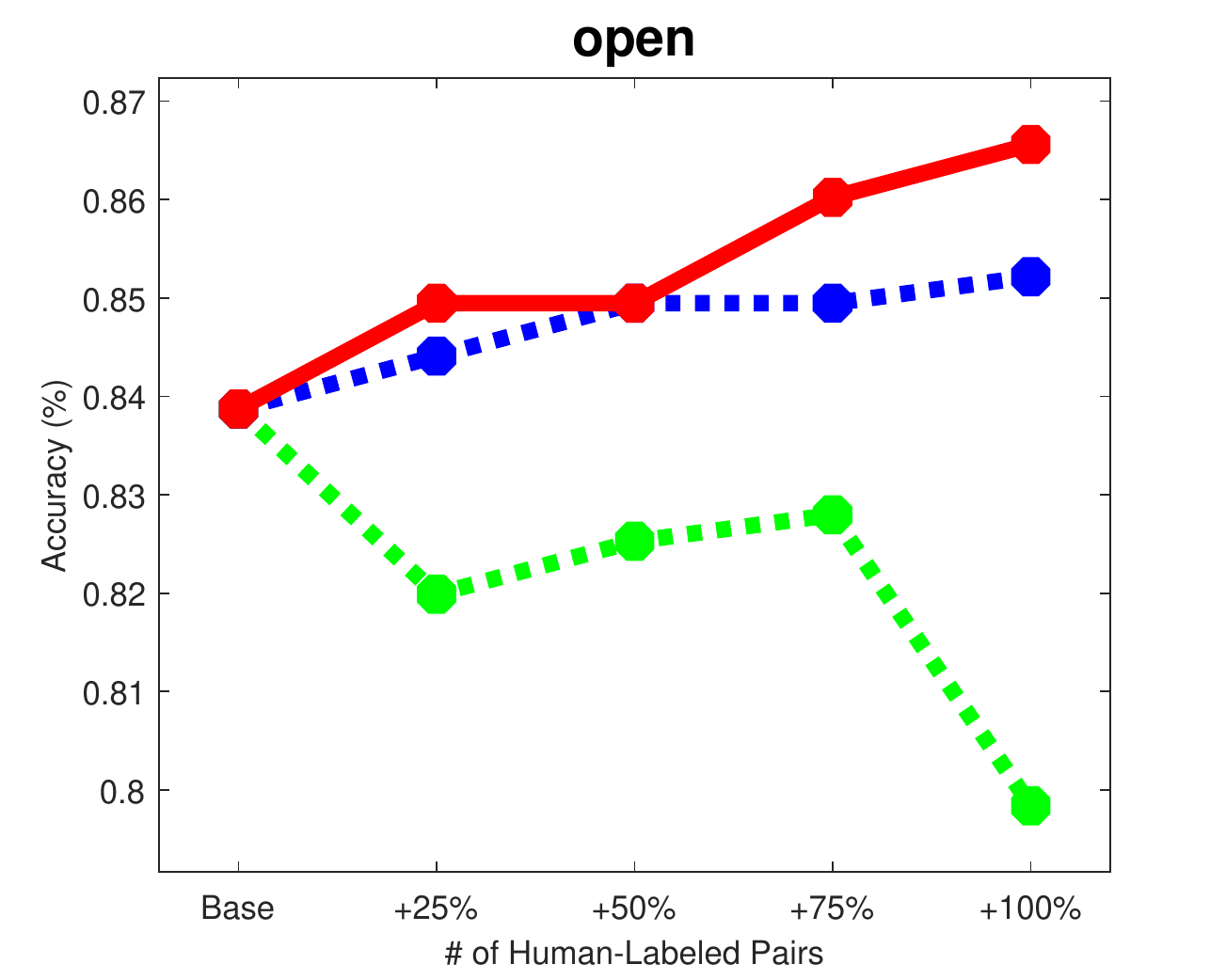}
    \end{subfigure}%
    \caption{Individual active learning curves for the shoes attributes.}
    \label{fig:gain_shoes}
\end{figure*}

\begin{figure*}
\centering
    \begin{subfigure}[t]{.19\textwidth}
        \includegraphics[width=\textwidth]{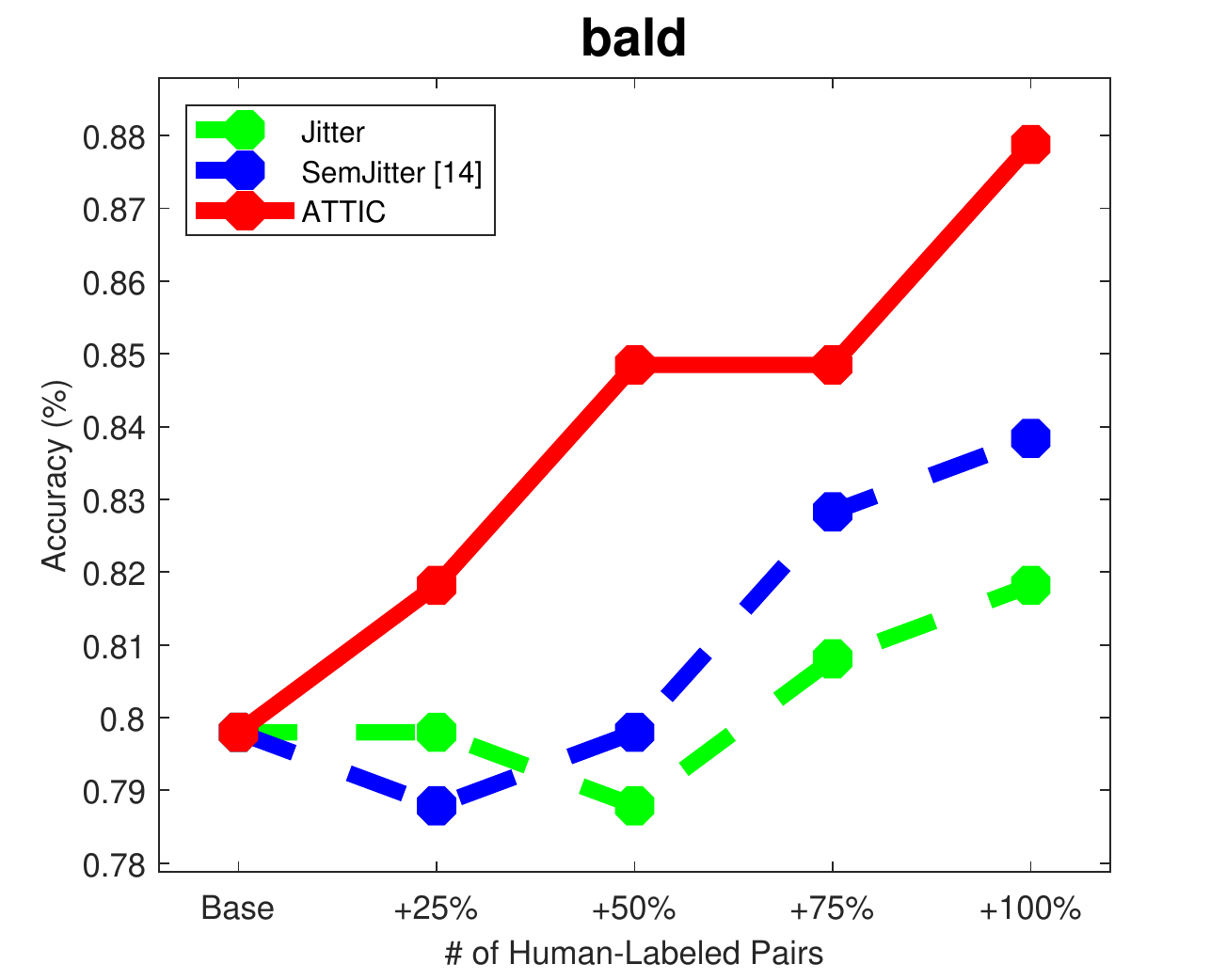}
    \end{subfigure}%
    \begin{subfigure}[t]{.19\textwidth}
        \includegraphics[width=\textwidth]{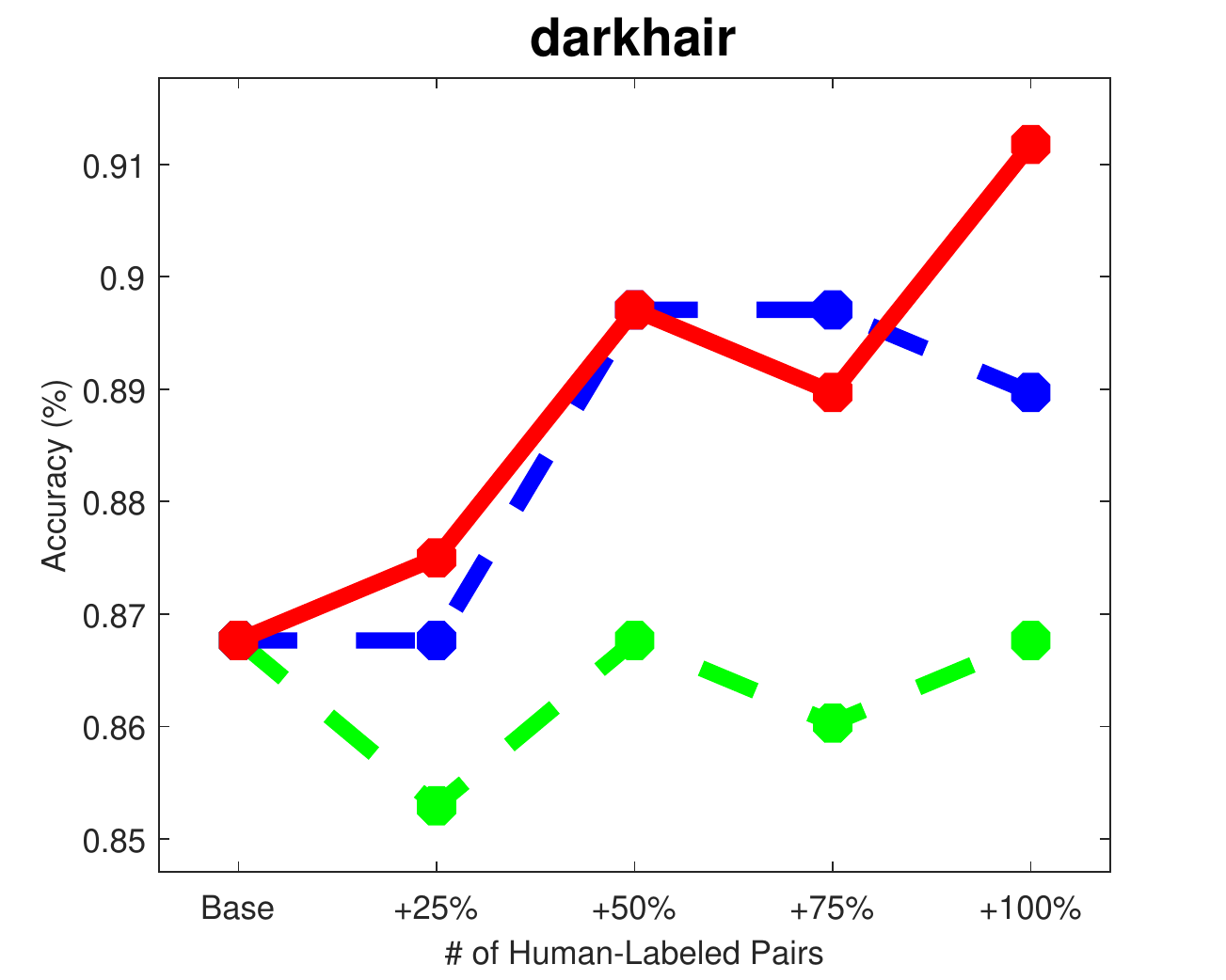}
    \end{subfigure}%
    \begin{subfigure}[t]{.19\textwidth}
        \includegraphics[width=\textwidth]{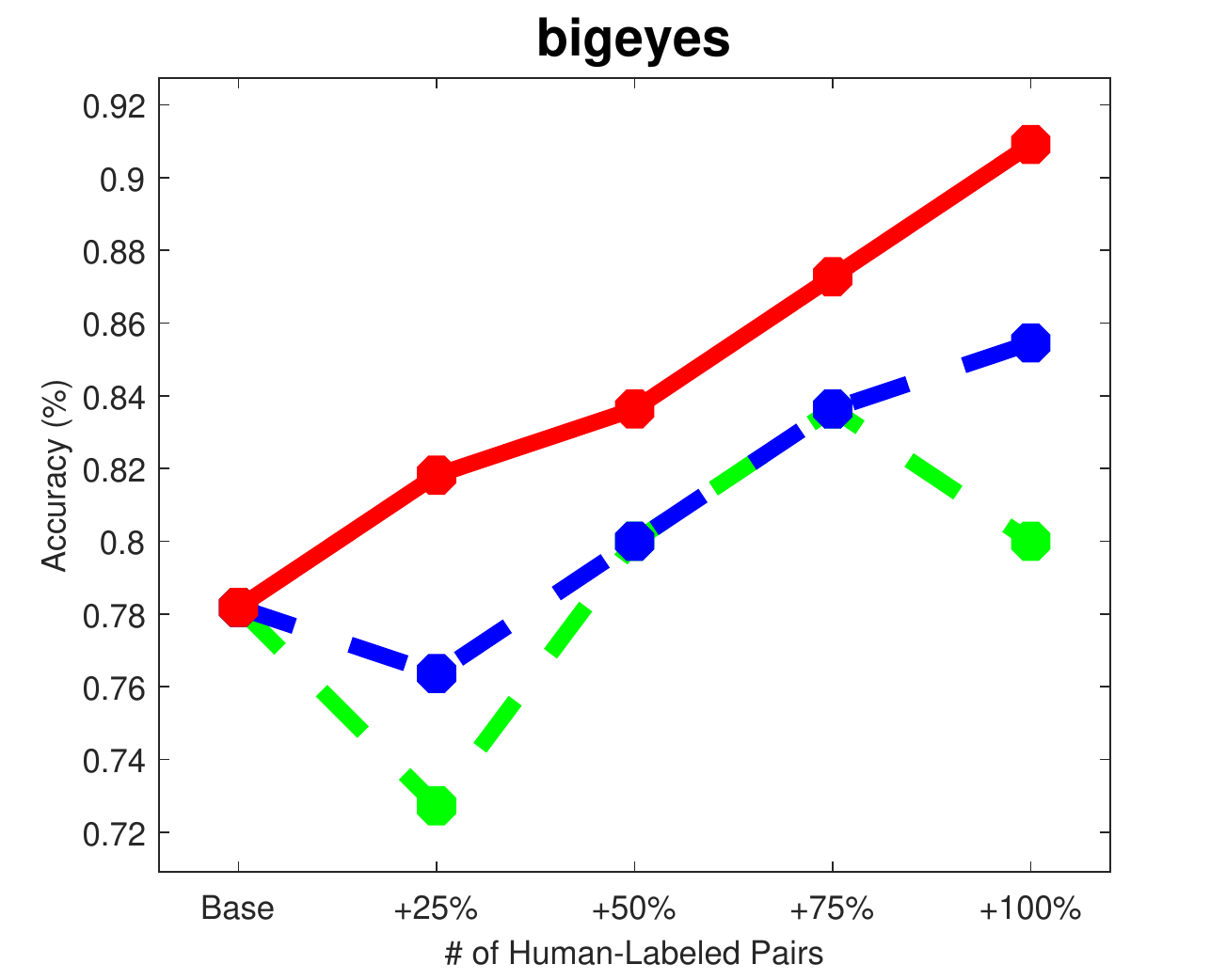}
    \end{subfigure}%
    \begin{subfigure}[t]{.19\textwidth}
        \includegraphics[width=\textwidth]{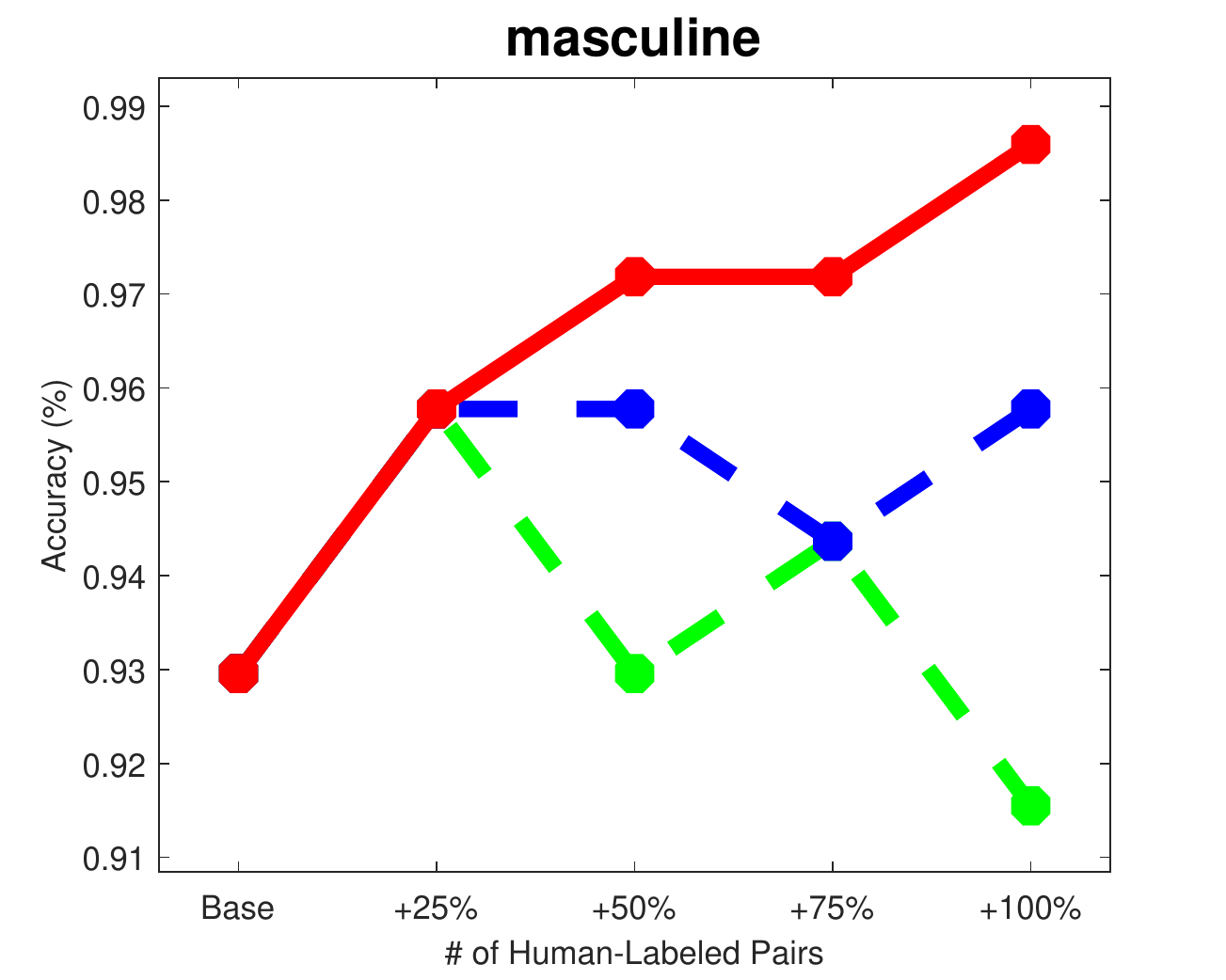}
    \end{subfigure}%
    \begin{subfigure}[t]{.19\textwidth}
        \includegraphics[width=\textwidth]{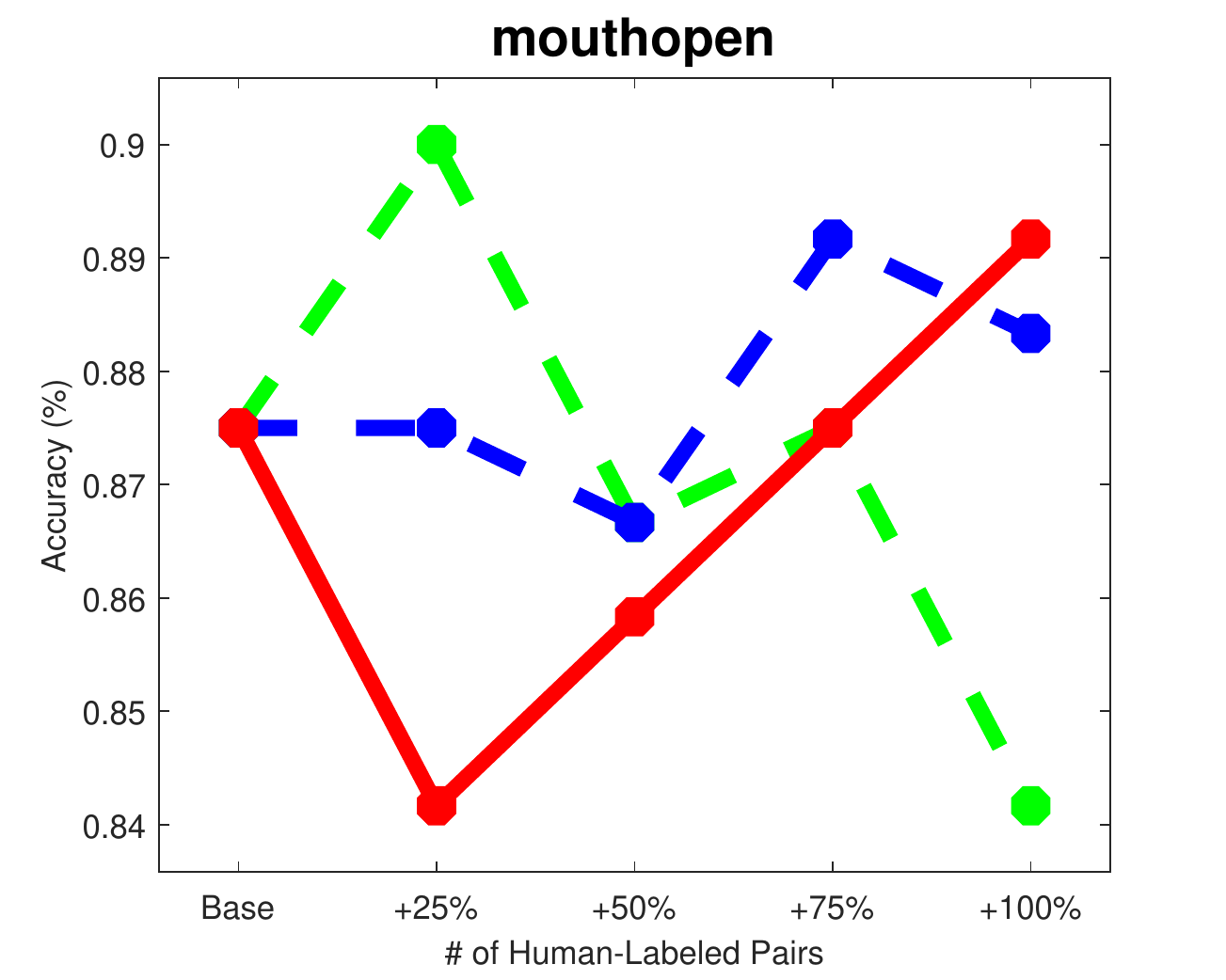}
    \end{subfigure}
    \begin{subfigure}[t]{.19\textwidth}
        \includegraphics[width=\textwidth]{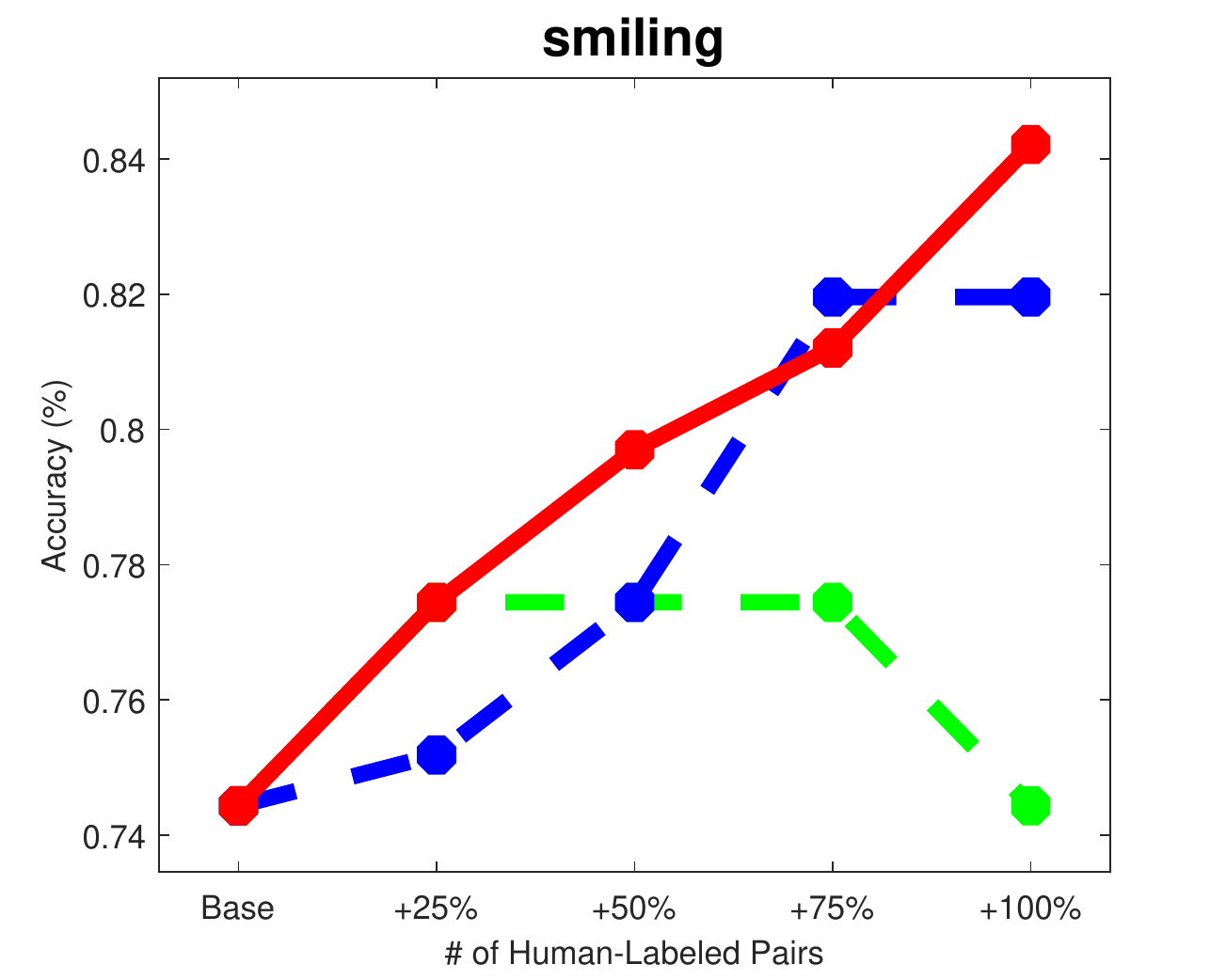}
    \end{subfigure}%
    \begin{subfigure}[t]{.19\textwidth}
        \includegraphics[width=\textwidth]{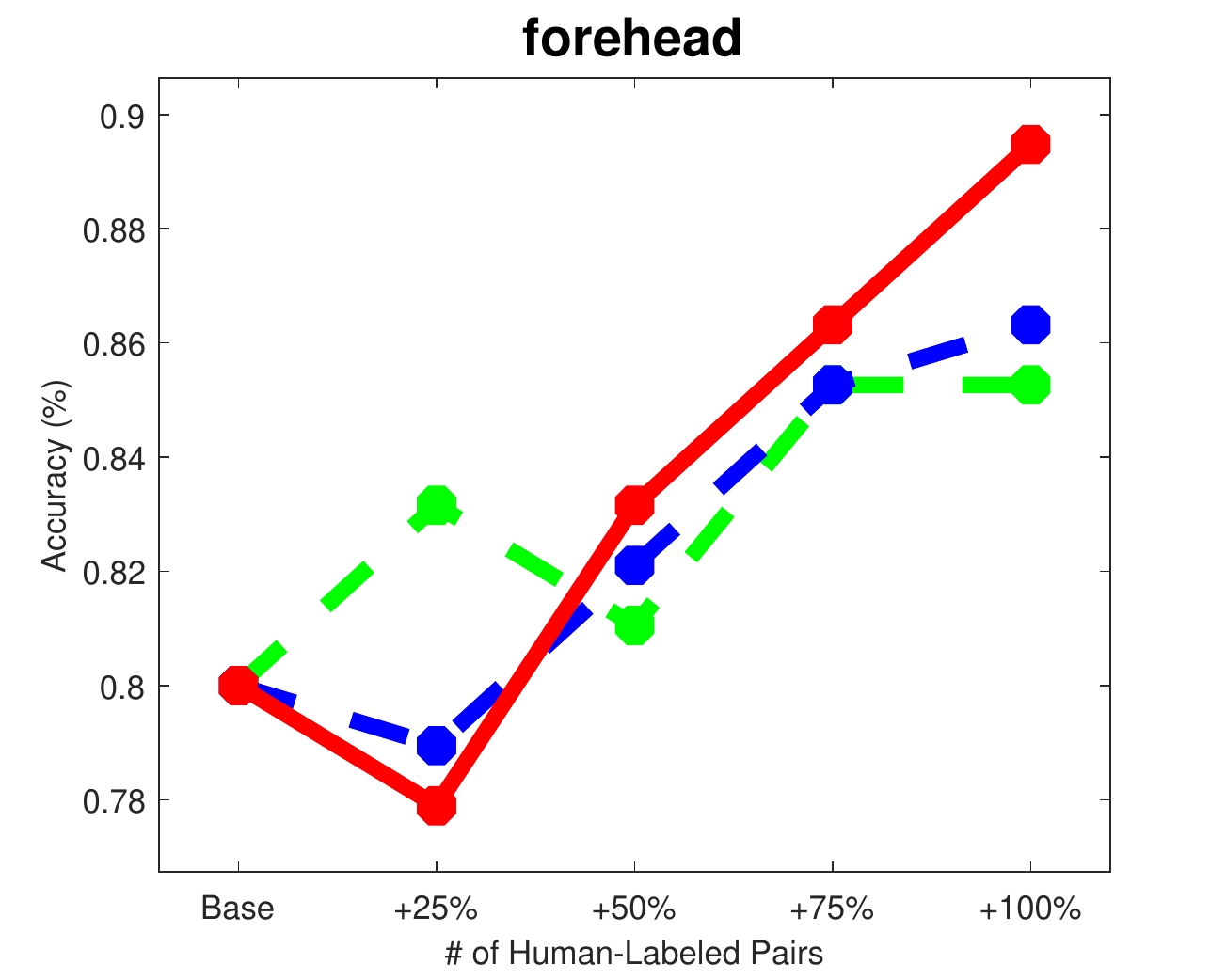}
    \end{subfigure}%
    \begin{subfigure}[t]{.19\textwidth}
        \includegraphics[width=\textwidth]{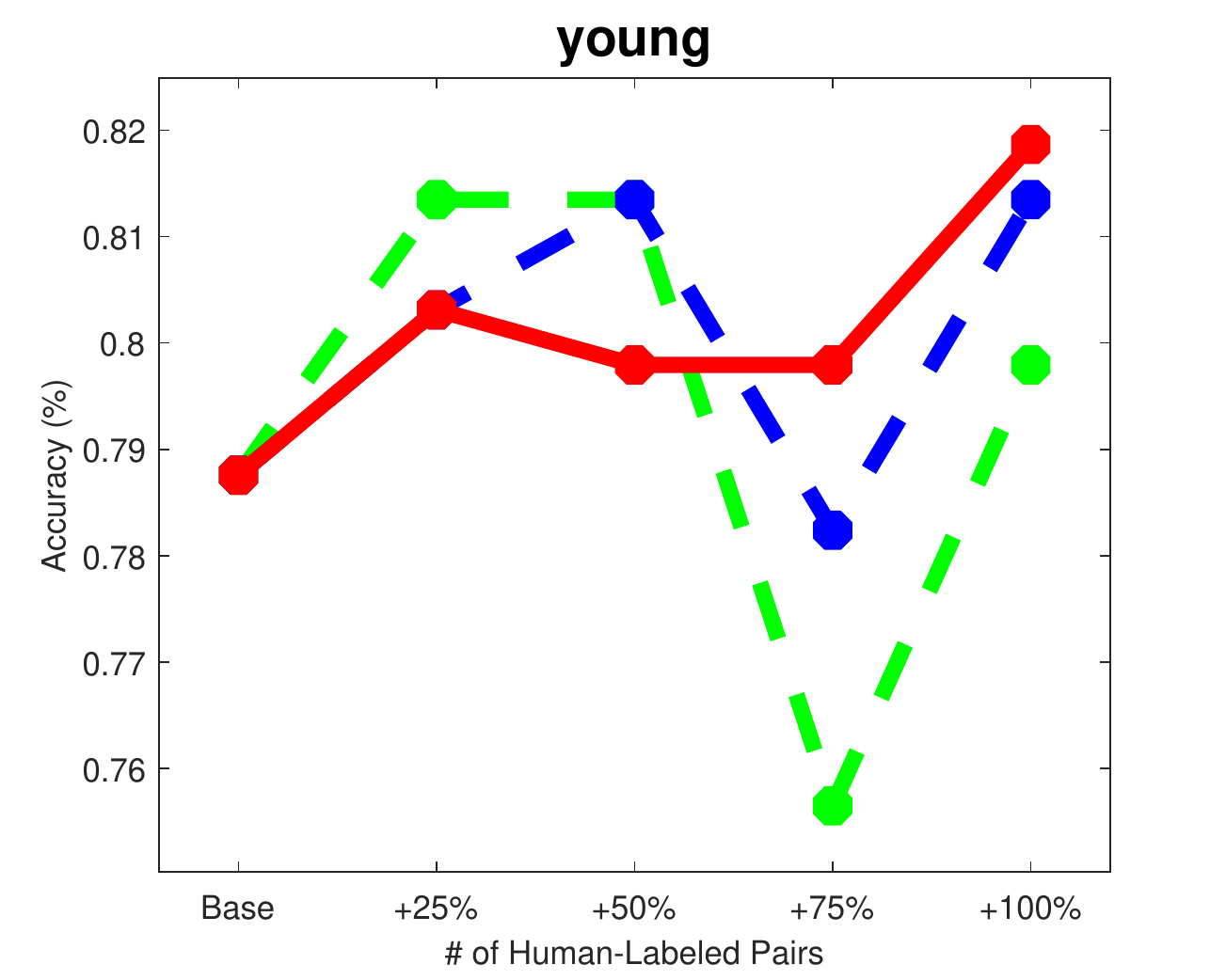}
    \end{subfigure}%
    \caption{Individual active learning curves for the face attributes.}
    \label{fig:gain_faces}
\end{figure*}


\begin{figure*}[t]
\centering
    \includegraphics[width=.6\textwidth]{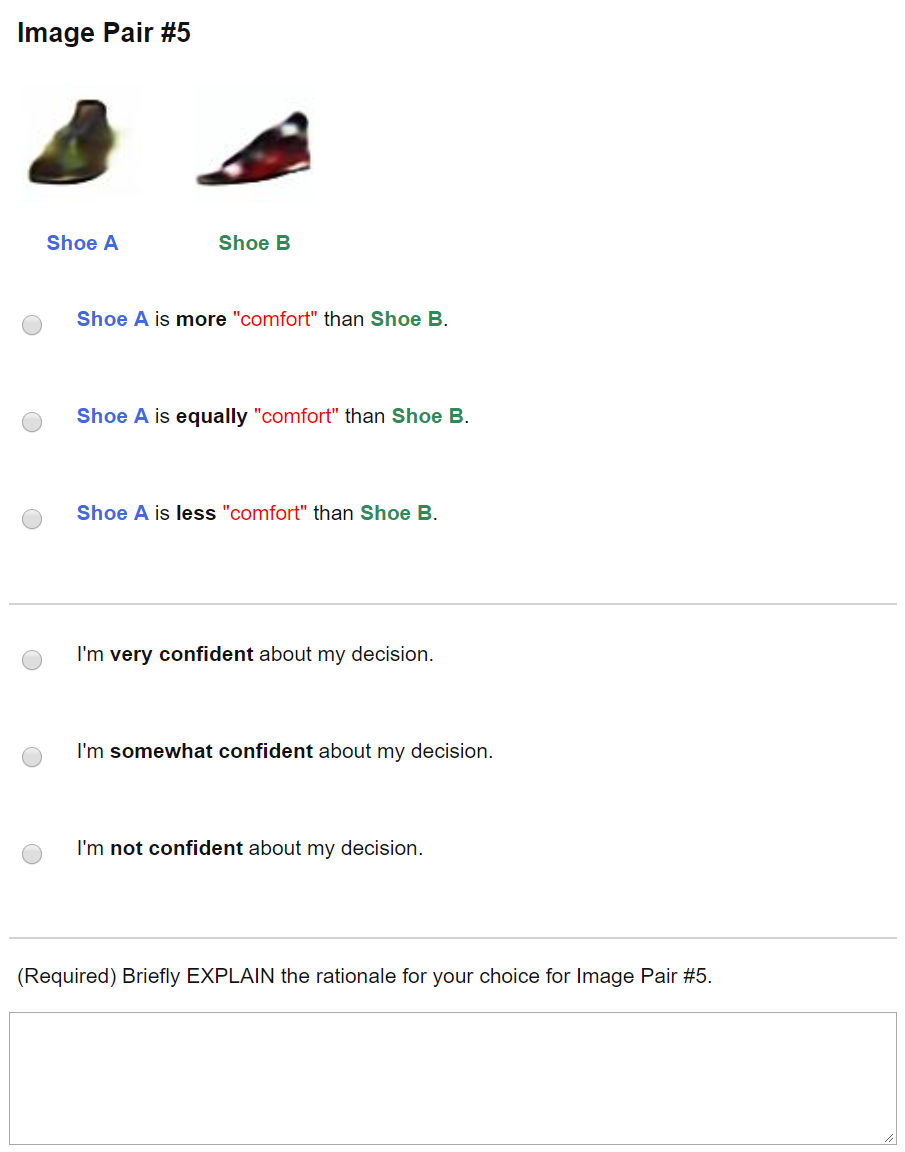}
    \caption{Example of a single task within a HIT.}
    \label{fig:mturk_hits}
\end{figure*}

\section{Appendix}
\label{sec:appendix}

\vspace*{0.1in}
\subsection{Active Training Individual Plots}
\vspace*{0.05in}

In Figure~\ref{fig:abatch_gain} of the main text, we present the gain curves of our active training experiment over the Real baseline, averaged over all attributes.  Here, we show the individual gain curves for each attribute from both datasets.  Figure~\ref{fig:gain_shoes} and \ref{fig:gain_faces} represent the shoes and face attributes, respectively.  Our approach learns the fastest for almost every attribute.

\subsection{Active Labels from Human Annotators}
\vspace*{0.05in}

In Figure~\ref{fig:mturk_hits}, we show the interface we used to collect labels from human annotators on MTurk for the actively synthesized training images.  In addition to the relative decision, we also instruct the workers to indicate their level of confidence with their decision.  Image pairs with low overall confidence and/or low agreement among workers are pruned and not used in training.

\end{document}